\documentclass{article}

\usepackage[utf8]{inputenc} 
\usepackage[T1]{fontenc}    

\usepackage{microtype}

\usepackage[inline]{enumitem}

\usepackage[normalem]{ulem}

\usepackage[draft=false,hidelinks]{hyperref}
\usepackage[numbered]{bookmark}  
\usepackage{parskip}

\usepackage[numbers,compress]{natbib}
\usepackage{doi}

\usepackage[nottoc, section]{tocbibind}

\usepackage[dvipsnames]{xcolor}

\usepackage[verbose=true,letterpaper]{geometry}

\usepackage[final]{graphicx}

\graphicspath{%
  {figures/}%
}

\usepackage{tikz}

\usetikzlibrary{%
  calc,%
  positioning,%
  arrows.meta,%
}

\usepackage{booktabs}
\usepackage{multirow}
\usepackage{wrapfig}

\usepackage{algorithm}
\usepackage[indLines=true]{algpseudocodex}

\renewcommand*\cite[1]{\citep{#1}}

\usepackage{subcaption}

\usepackage[dvipsnames]{xcolor}
\usepackage[english]{babel}
\usepackage{amsmath,amssymb}
\usepackage{bm}
\usepackage{multirow}
\usepackage{diagbox}

\DeclareBoldMathCommand\vx{x}
\DeclareBoldMathCommand\vy{y}
\DeclareBoldMathCommand\vtheta{\theta}
\DeclareBoldMathCommand\veps{\epsilon}
\DeclareBoldMathCommand\vzero{0}

\newcommand{\idmat}{\mathrm{1}}

\newcommand{\mat}[1]{\ensuremath{\mathrm{#1}}}
\newcommand{\normal}[2]{\ensuremath{\mathcal{N}\left(#1, \,#2\right)}}

\newcommand{\dif}{\ensuremath{\mathrm{d}}}
\newcommand{\difftime}{\ensuremath{\tau}}

\newcommand{\degree}[1]{^{\circ}\,\mathrm{#1}}

\newcommand{\labelize}[1]{\textsf{\textbf{#1}}}

\usepackage[most]{tcolorbox}  
\usepackage{enumitem}         

\usepackage[
  capitalize,
  nameinlink,
  noabbrev,
]{cleveref}

\title{A Generative Framework for Probabilistic, Spatiotemporally Coherent Downscaling of Climate Simulation}
\author{
  Jonathan Schmidt$^{*, 1, 2}$, Luca Schmidt$^{1, 2}$, Felix M. Strnad$^{1, 2}$,\\ Nicole Ludwig$^{1, 2}$, Philipp Hennig$^{1, 2}$\\[2ex]
  \normalsize
  \begin{tabular}{rl}
    $^1$: & University of Tübingen, Tübingen, Germany \\
    $^2$: & Tübingen AI Center, Tübingen, Germany \\
    $^*$: & Corresponding author. Email: \texttt{jonathan.schmidt@uni-tuebingen.de}
  \end{tabular}\\[2ex]
}
\date{}

\begin{document}




\maketitle

\bigskip

\begin{tcolorbox}[colback=red!5!white]
  \centering
  Please find the published and edited version of this manuscript here:\\[1em]
  \url{https://www.nature.com/articles/s41612-025-01157-y}
\end{tcolorbox}

\noindent\rule{\linewidth}{1pt}
\paragraph{Abstract} Local climate information is crucial for impact assessment and decision-making, yet coarse global climate simulations cannot capture small-scale phenomena. Current statistical downscaling methods infer these phenomena as temporally decoupled spatial patches. However, to preserve physical properties, estimating spatio-temporally coherent high-resolution weather dynamics for multiple variables across long time horizons is crucial.
We present a novel generative framework that uses a score-based diffusion model trained on high-resolution reanalysis data to capture the statistical properties of local weather dynamics. After training, we condition on coarse climate model data to generate weather patterns consistent with the aggregate information.
As this predictive task is inherently uncertain, we leverage the probabilistic nature of diffusion models and sample multiple trajectories.
We evaluate our approach with high-resolution reanalysis information before applying it to the climate model downscaling task. We then demonstrate that the model generates spatially and temporally coherent weather dynamics that align with global climate output.

\noindent\rule{\linewidth}{1pt}



\section*{Introduction}

\noindent Numerical weather and climate simulations based on discretized solutions of the Navier-Stokes equations are fundamental to understanding large-scale weather patterns, climate variability, and climate change.
State-of-the-art numerical weather prediction (NWP) models, which primarily focus on atmospheric processes, can accurately resolve small-scale dynamics within the Earth system, providing fine-scale spatial and temporal weather patterns at resolutions on the order of kilometers \citep{Bauer2015}.
However, the substantial computational resources required for these models render them impractical for simulating the extended time scales of multiple years and decades necessary to assess climatic changes.
Moreover, even with substantial computational investment, global high-resolution models can still exhibit systematic biases and may fail to accurately reproduce observed climatic trends \cite{moon2024globalhresm}.
In contrast, earth system models (ESMs), such as those included in the CMIP6 project \citep{Eyring2016}, incorporate a broader range of processes---including atmospheric, oceanic, and biogeochemical interactions---while operating on coarser spatial scales.
Typical grid resolutions for ESMs are approximately $1^{\circ}$, equivalent to around 100 km. This coarse resolution limits the ability of ESMs to fully capture small-scale processes.
Key processes necessary to assess regional impact, for example, on wind turbines---such as local wind turbulence---occur at spatial and temporal scales that are too fine to be explicitly resolved in ESMs.
Consequently, ESM data cannot be directly employed to evaluate changes at fine spatial scales, limiting their utility for localized impact assessment and decision-making.

Downscaling aims to provide regional climate information by estimating small-scale processes from coarse simulations of global models.
Existing approaches to bridge this scale gap can be categorized into dynamical and statistical downscaling \cite{maraun2018dsbook}.
Dynamical downscaling employs high-resolution regional climate models (RCMs) that are nested within coarser global climate models (GCMs) to provide detailed projections for specific regions \citep{Laprise2008, Tapiador2020}.
Biases from the driving global models can be inherited by the RCMs, potentially limiting the accuracy of the downscaled results \citep{Risser2024}.
Additionally, the high computational cost of running RCMs restricts their use primarily to regional studies.
Statistical downscaling methods use regression or weather generators \citep{wilks1999weathergenerator}, and, more recently, emulators based on machine learning (ML) techniques.
In statistical downscaling, a functional or statistical relationship between large-scale climate variables (from GCMs) and local observations is established.
Based on this relationship, statistical methods infer the local information from the coarse simulation, with ML-based approaches aiming to capture the mapping from coarse to fine-scale climate by learning the relationships from data \cite{gmd-13-2109-2020,BanoMedina2021,Balmaceda-Huarte2024,BABAOUSMAIL2021105614}. These models are usually computationally less expensive. However, not explicitly encoding physical laws can make them less robust in regions of low data density, risking physical inconsistencies.

\begin{figure}[htbp]
  \centering
  \includegraphics[width=\linewidth]{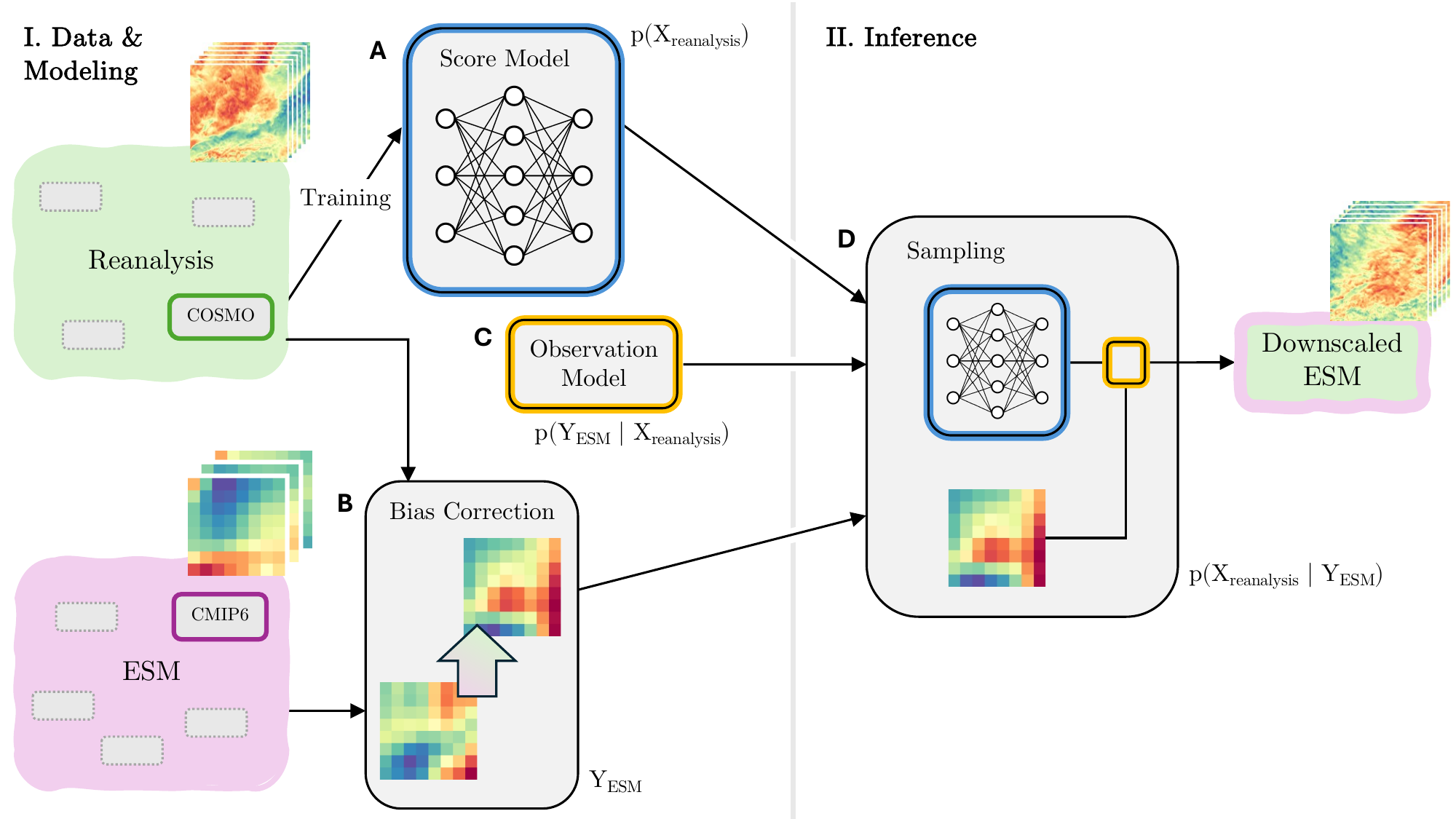}
  \caption{\textbf{Probabilistic pipeline for spatiotemporal downscaling of multiple variables.}
    This work introduces a probabilistic downscaling framework that jointly predicts fine-scale and spatiotemporally consistent time series for multiple variables from coarse ESM simulations.
    This schematic outlines the framework. Only one exemplary variable is shown for visual clarity.
    \labelize{A}: A score model is trained on sequences of reanalysis data. This is the centerpiece of the diffusion model that learns to reproduce the fine-scale spatial and temporal patterns. Note that ESM simulations are not part of the training process. \labelize{B}: From any ESM (e.g. CMIP6), select and pre-process an ensemble run (e.g. MPI-HR) for downscaling. A bias-correction step that mitigates distributional deviations between climate output and reanalysis data can be applied as a pre-processing step.
    \labelize{C}: This part establishes a relationship between coarse climate simulations and the (fine-scale) output space of the model. The observation model defines how the observed quantity ($Y_\mathrm{ESM}$) is generated---or observed---from the inferred quantity ($X_\mathrm{reanalysis}$).
    The observation model is key to impose a constraint onto the generative model such that its samples adhere to the established relationship.
    \labelize{D}: The model generates time series that preserve the statistics of the coarse climate input. During the generative process, the trained score model (\labelize{A}) is conditioned, i.e., the predictions are informed by the conditioning information (\labelize{B}), such that they adhere to the relationship established by the observation model (\labelize{C}).
  }
  \label{fig:pipeline-sketch}
\end{figure}

One challenge that comes with downscaling arises from the fact that numerical models are inherently imperfect representations of the climate system.
While ESMs are designed to generate accurate multi-decadal summary statistics, locally, the simulations differ from historical observational datasets.
Climate models contain inaccuracies in parameterization, simplified process representations, uncertainties in the initial state of the system \cite{smith2019robust}.
Even in the hypothetical case of perfect models, i.e., without epistemic uncertainty, forecasts are not deterministic due to the chaotic nature of the atmosphere.
These variations lead to substantial discrepancies between models.
Choosing a model, therefore, becomes a critical factor \cite{doblasreyes2021linking,morelli2024climate}.
Aside from other climate model biases \cite{chen2021framing}, internal variability of the climate system leads to differences between projected and observed climate \cite{jain2023importance}.
Individual climate simulations thus represent only one possible realization of the system with substantial uncertainty remaining.
Due to this un-pairedness of ESM outputs and observational data, using supervised ML approaches, which rely on consistent simulation--observation pairs, remains challenging \cite{Hess2022physically,Aich2024}.
Generative models have recently emerged as a promising solution.
This model class is characterized by learning a representation of the training data distribution that allows the generation of novel samples.
As self-supervised learning techniques, these models circumvent the need for data--label pairs by working solely on the target (output) distribution.
Additionally, through the variability among generated samples, generative models provide structured uncertainty, which the ill-posed nature of most inference problems entails.
In particular, diffusion models (DMs) \citep{ddpm2020ho,sdedm2021song,ddim2021song,sda2023rozet} have demonstrated superior performance over earlier approaches such as variational auto-encoders (VAEs) \cite{vae2014kingma}, generative adversarial networks (GANs) \citep{gan2014goodfellow}, and normalizing flows \citep{Dinh2014}, particularly for structured data and image generation tasks \cite{guidance2021dhariwal,Cao2022, Koo2023}.

In this work, we build on the score-based data assimilation (SDA) framework by \citet{sda2023rozet,sdaqg2023rozet} to phrase downscaling as a Bayesian-inference problem with a generative prior model.
By construction, our model performs \emph{joint} spatial and temporal downscaling on multiple variables, introducing stochasticity solely between samples.
The predictions are thus coherent across the entire state space, avoiding sampling-induced inconsistencies between time steps and between interrelated atmospheric variables.
Furthermore, the model training is separated from the task-specific inference (Fig.~\ref{fig:pipeline-sketch}~\textbf{I} \& \textbf{II}), which makes the model flexible with respect to its input and the statistical relationship between input and output---allowing, for instance, downscaling climate simulations of varying spatiotemporal resolutions without retraining the model.
The components of the algorithmic pipeline are outlined in \cref{fig:pipeline-sketch}:
\begin{itemize}
  \item[\labelize{A}] The score model%
    , which is the centerpiece of the generative diffusion model, is trained on high-resolution reanalysis data. The score-based diffusion model provides a statistical representation of local weather dynamics.
    This \textit{"prior"} model gives access to samples from the distribution $p(X_\mathrm{reanalysis})$, which represents our prior concept of the output space. Here, the output space includes the spatial region, the target resolution, and learned dynamics patterns for a set of selected atmospheric variables.
  \item[\labelize{B}] The coarse ESM input $Y_\mathrm{ESM}$ is pre-processed; in particular, a bias-correction procedure can align the simulation with the reanalysis data in terms of its value distribution.
  \item[\labelize{C}] The \textit{"observation model"} $p(Y_\mathrm{ESM} \mid X_\mathrm{reanalysis})$ %
    establishes the functional or statistical relationship between coarse climate output and fine reanalysis data. This assumes the ESM output to be a perfect prognosis \cite{klein1959perfectprog,maraun2010perfectprog}, based on which local climate is estimated.
    In the context of inverse problems, the observation model is often referred to as the "forward model", as it models how the partially observed quantity ($Y_\mathrm{ESM}$) arises from the latent quantity of interest ($X_\mathrm{reanalysis}$)---usually by removing information.
  \item[\labelize{D}] Accordingly, the inverse problem aims to predict the (much harder, underspecified) opposite direction: estimating the \textit{"posterior"} $p(X_\mathrm{reanalysis} \mid Y_\mathrm{ESM})$ requires adding information to the incomplete observations by conditioning the prior.
    In our case, the trained score model is coupled with the observation model to generate samples from the posterior distribution.
\end{itemize}

Perspectively, the algorithmic framework is versatile and flexible enough to be considered beyond downscaling.
In the spirit of foundation models \citep{Lessig2023,Lam2023, Price2025, Yang2025}, the trained generative model can be combined with other observation models in a similar zero-shot manner to solve various inference tasks on the target output space.
Unlike approaches that require training the model directly on the conditioning information, the presented framework allows the formulation of explicit---and varying---functional or statistical relationships between observations and predictions.

On a series of experiments, we demonstrate that the proposed model generates coherent time series of regional climate that are aligned with coarse input.
The model predicts local weather dynamics, including extreme events such as winter storms.
Sampling from the posterior distribution enables the generation of multiple weather trajectories crucial for assessing the internal variability and uncertainty of the downscaling problem, providing a comprehensive understanding of future weather scenarios.
We include a simple quantile-mapping procedure as a pre-processing step to the ESM simulations.
Other than that, the ESM simulations are taken as a perfect predictor and the mismatch to the observational data is not accounted for.
We evaluate our downscaling framework on two different GCMs from CMIP6 in order to show that different mismatches between reanalysis and the respective ESM distributions remain.
We then demonstrate that our model maintains the coarse, global properties of the respective ESM input while inferring local fine-scale weather patterns.


\section*{Results}\label{sec:results}

The presented downscaling pipeline assumes a statistical relationship between a coarse numerical model prediction and a fine-scale reanalysis product \cite{klein1959perfectprog,maraun2010perfectprog}.
We begin by evaluating the methodological framework in an artificial setup, in which coarsened reanalysis data serve as the perfect prediction \cite{maraun2015VALUE} and surrogate the ESM simulations.
The coarse data is obtained from spatial area averages and by selecting a subset of the time steps.
Concretely, the observation model establishes the following relationship between the downscaled predictions $X$ and the coarse input $Y$:
\begin{equation}\label{eq:stat-relationship-obsmodel}
  Y^{(i, j)}_{t} = \frac{1}{\lvert\overline{i}\rvert \cdot \lvert\overline{j}\rvert} \cdot \sum_{(i', j') \in \overline{i} \times \overline{j}} X^{(i',j')}_t,
\end{equation}
where $(i, j)$ is a single point on the coarse spatial grid, which represents an area of multiple points $\overline{i} \times \overline{j}$ on the fine grid.
For example, $\overline{i}$/$\overline{j}$ can be a local neighborhood around $i$/$j$ in the longitude/latitude dimension.
The observation model in \cref{eq:stat-relationship-obsmodel} assumes that the coarse information is provided as a snapshot at time points $t$.
In our experiments, one coarse-grid point $(i, j)$ encompasses a $16 \times 16$-area on the fine grid, i.e., $\lvert\overline{i}\rvert \cdot \lvert\overline{j}\rvert = 16^2 = 256$.
Furthermore, the temporal resolution of the coarse grid is six hours, whereas the fine output time grid is resolved hourly.
This setting enables a direct pairing between climate output and reanalysis data, which is not given in reality \cite{maraun2015VALUE}, allowing for evaluating the model's predictive performance and uncertainty quantification by comparing it to a ground truth.
Having established the validity of the downscaling method on this artificial setup, we will then proceed to downscale climate output, assuming two different ESM simulations from the CMIP6 project to be the perfect predictors.

\subsection*{Evaluation of predictive distribution and uncertainty calibration}

\begin{figure}[htbp]
  \centering
  \includegraphics[width=\linewidth]{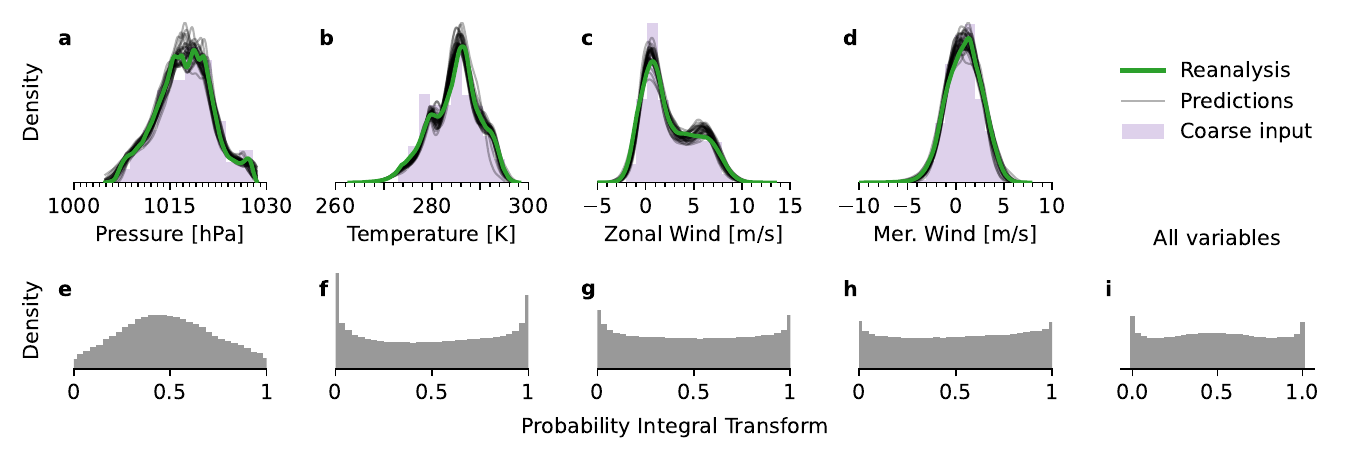}
  \caption{%
    \textbf{Comparison of value distributions: reanalysis data, coarse input, predictions.}
    For a time range of 49 hours, this plot shows that the 1-hourly local predictions, which the model predicted from the coarse 6-hourly input, resemble the reanalysis data closely in distribution.
    \emph{Top row:} Kernel-density estimations for value distributions of reanalysis data (green) and 30 predictions (black).
    The prediction model was conditioned on coarse inputs (purple).
    The predicted samples align with the reanalysis data distribution, which is fully covered by the predictive uncertainty.
    \emph{Bottom row:} The probability integral transform (PIT) demonstrates the uncertainty calibration of the model: for each variable separately (\labelize{e}--\labelize{h}) and overall (\labelize{i}).
    A PIT distribution that resembles a standard uniform distribution indicates that the reanalysis data and predictions likely come from the same distribution.
  }
  \label{fig:on-model-value-histogram}
\end{figure}

We first evaluate the basic capability of the model to adhere to the established statistical relationship between coarse and fine predictor (cf.~\cref{eq:stat-relationship-obsmodel}) and to preserve the coarse-data value distribution in its predictions.
Aggregating the values over temporal and spatial domains, we find that---per variable---the estimated densities of the predictions each align with the reanalysis data.
The uncertainty induced by the sample spread covers the reanalysis distribution both near the modes and in regions of low and high quantiles.
In particular, the probabilistic model introduces no systematic biases, like distribution shift, a tendency towards over- or under-predicting values, or mismatch in the tail regions.
Instead, each sampled prediction captures the spread of the data distribution.
\Cref{fig:on-model-value-histogram} \labelize{a}--\labelize{d} visualizes these findings using density estimations of the respective value distributions.

To assess the calibration of the predictive uncertainty, we compute the probability integral transform (PIT) \cite{dawid1984pit} for all values aggregated and for each variable separately.
If the resulting distribution (\cref{fig:on-model-value-histogram}~\labelize{e}--\labelize{i}) resembles a standard uniform distribution on the interval $[0, 1]$, it is likely that the samples and the reanalysis data come from the same underlying distribution.
We find that both wind-speed components (\cref{fig:on-model-value-histogram}~\labelize{g},\labelize{h}) are well calibrated.
The PIT reveals that the predictions for mean sea-level pressure (\cref{fig:on-model-value-histogram}~\labelize{e}) are slightly underconfident, over-predicting extreme values. For surface temperature (\cref{fig:on-model-value-histogram}~\labelize{f}), the opposite is the case: the predicted distribution is slightly too narrow and under-predicts tail events.
Taken the joint distribution of variables together (\cref{fig:on-model-value-histogram}~\labelize{i}), the model is well calibrated.


\subsection*{Predicting local dynamics from coarse information}

\begin{figure}[htbp]
  \centering
  \includegraphics[width=\linewidth]{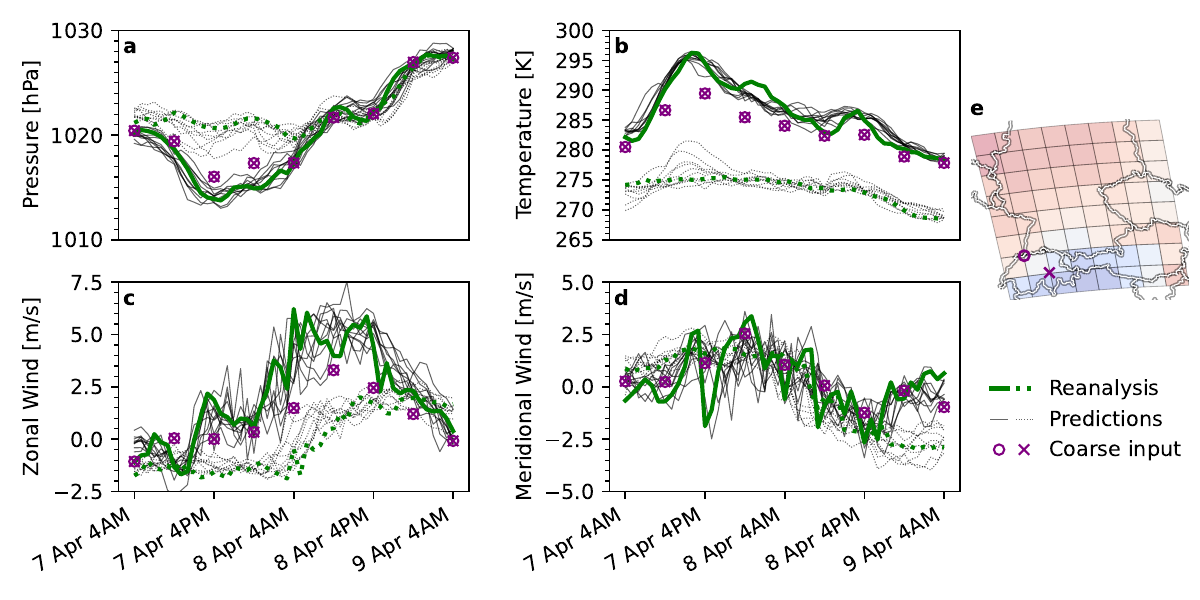}
  \caption{
    \textbf{Differences in local dynamics are inferred from coarse-grained observations.}
    This plot compares time series at two locations over 49 hours between reanalysis data (green), 30 predictions (black), and coarse input (purple) and shows that the model can accurately extract local dynamics from shared coarse information. Both fine-grid locations were selected to share a single point on the coarse grid (\labelize{e}).
    Each plot (\labelize{a}--\labelize{d}) shows one variable. Reanalysis and predicted weather trajectories are shown in solid and dotted lines for both locations, respectively.
    The conditioning information (purple circles and crosses) is a single value every six hours that is shared by both locations (\labelize{e}), at which the local weather dynamics are inferred.
    The predicted time series aligns with the reanalysis data at both locations.
    In particular, the uncertainty obtained through sampling multiple predictions covers the reanalysis data, and the individual samples mirror the local weather trajectories in their respective temporal structure.
  }
  \label{fig:on-model-timeseries}
\end{figure}

We require our downscaling model to augment the scarce information contained in the coarse input by adding nontrivial, local weather patterns and predicting complex temporal and spatial dynamics.
As described above, in our experimental setup, a single scalar measurement informs a six-hour window of $16\times 16$ fine-grid locations, requiring the model to perform a mapping from a single node to $6\times 16 \times 16 = 1\,536$ nodes.
Hence, we need to assess whether the model sensibly incorporates prior knowledge, which it learned from data, in order to evaluate the plausibility of the added, generated information.

It is likely that---due to, for instance, varying environmental conditions---the dynamical patterns at distinct locations on the reanalysis grid differ substantially from each other, whereas the coarse-grid aggregation occludes these local variations, motivating the downscaling problem in the first place.
The present input-output-paired setup allows us to investigate the true local variations that are lost through aggregation by comparing the reanalysis data at two distant locations within the $16$-by-$16$ area that is encompassed by a single coarse-grid location.
Accordingly, we can compare these ground-truth variations with our model predictions to assess whether the model infers local dynamics that align with the reanalysis data.

In \cref{fig:on-model-timeseries}, we demonstrate that our model accurately predicts spatial and temporal variations in weather trajectories at two distinct locations, which share a single spatial observation.
For this experiment, we selected two distant fine-grid locations near the Alps, for which we can expect substantial local variations (see purple circle and cross in \cref{fig:on-model-timeseries}~\labelize{e}).
We visualize the time series for the reanalysis data at both locations (solid and dotted green line) and the corresponding downscaled model predictions (solid and dotted black lines) alongside the coarsened input (purple circle and cross), which share a single value at every 6-hourly step.
The visualization allows to identify the local variations that are occluded in the spatiotemporal aggregation by comparing the coarse observations (purple) to the fine ground-truth time series.
Our model accurately predicts the ground-truth dynamics at both locations from the coarse information.
Notably, the distinct features in temporal structure (e.g. smoothness and amplitude) of the predicted time series at both locations align with those found in the corresponding reanalysis data.
As an illustrative example, the zonal wind speeds (\cref{fig:on-model-timeseries} \labelize{c}) show substantial variability between the two locations---both in their values and their temporal structure. The model captures these variations in its predictions, highlighting its ability to recover local-scale variability lost in the coarse-resolution observations.
This is a promising indicator that the downscaling model has learned an accurate representation of local spatial and temporal patterns from the training data, which it blends into the coarse conditioning information to estimate coarsely informed local variations.


\subsection*{Spatiotemporally consistent weather trajectories: studying a winter storm}

\begin{figure}[htbp]
  \centering
  \includegraphics[width=\linewidth]{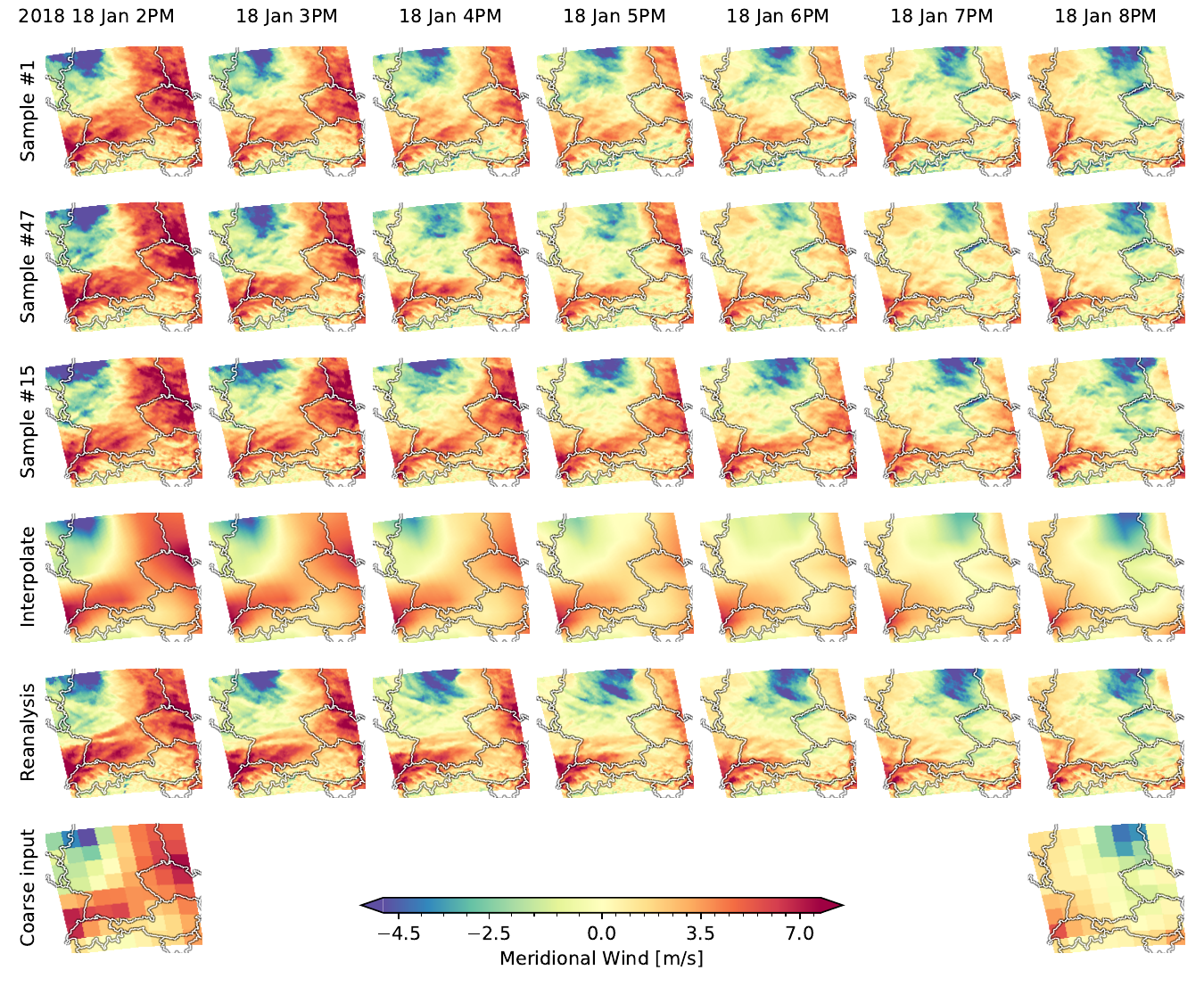}
  \caption{
    \textbf{Predicting high-resolution dynamics during a cyclone as an extreme event.}
    This plot shows meridional wind trajectories of three randomly selected model predictions (top three rows),
    spatiotemporal interpolation (fourth row), reanalysis data (fifth row), and coarse conditioning information (bottom row) during a cyclone ("Friederike", January 2018).
    The sign of the wind speed value defines its direction: negative values go southward, and positive values go northward.
    Time progresses from left to right hourly, starting 2018 January 18 at 02:00 PM and ending the same day at 08:00 PM.
    Between the first (2:00 PM) and the last (8:00 PM) visualized time point, no information is provided to the model.
    The top three rows show how different samples add missing spatial and temporal information, while the interpolation (fourth row) can only spread out the existing, coarse information.
    Each individual generated trajectory aligns visually with the coarse input, while the variation among the samples captures the uncertainty associated with the inference problem.
    Notably, the model does not introduce implausible ``jumps'' from one time step to the next but interpolates with spatially and temporally consistent dynamics.
  }
  \label{fig:on-model-storm}
\end{figure}

Statistical downscaling becomes particularly challenging during extreme events, which are rare and fall into the tails of the training distribution. These events often involve highly complex dynamics.
However, extreme events are particularly interesting, as they often have the most significant societal and environmental impacts.
As \cref{fig:on-model-value-histogram} already demonstrated, the downscaling model is capable of accurately predicting high quantiles.
Here, we supplement the evaluation of the aggregated value distributions with a qualitative assessment of the spatiotemporal structure of the downscaled predictions during a winter storm.
To this end, we consider the time range in which cyclone "Friederike" approached central Europe (including the modeled spatial region) from the west around January 18, 2018.
Aside from the time period, the experimental setup remains unaltered from the above sections.
\Cref{fig:on-model-storm} exemplary shows the meridional wind speeds during the event.
The visualization compares three randomly selected model predictions, a spatiotemporal interpolation of the coarse input, the ground-truth reanalysis data, and the coarse input.
The model predictions are coherent in space and time and add different local variations to the coarse observations.
A comparison to the spatiotemporal interpolation (\cref{fig:on-model-storm}, fourth row) highlights the capability of the downscaling model to predict nontrivial fine-scale patterns.
Supplementary Figure 9 visualizes anomalies---differences between the interpolation and a) downscaled predictions and b) reanalysis data.
This visualization reveals the spatiotemporal patterns, which are lost by aggregating the reanalysis data, and allows a comparison to the disaggregated fine-scale structure predicted by our downscaling model.
Additionally, Supplementary Figure 2 demonstrates a close match of power spectral densities between reanalysis data and model predictions, providing a more quantitative argument.

\subsection*{Downscaling ESM simulations}

\begin{figure}
  \centering
  \includegraphics[width=\linewidth]{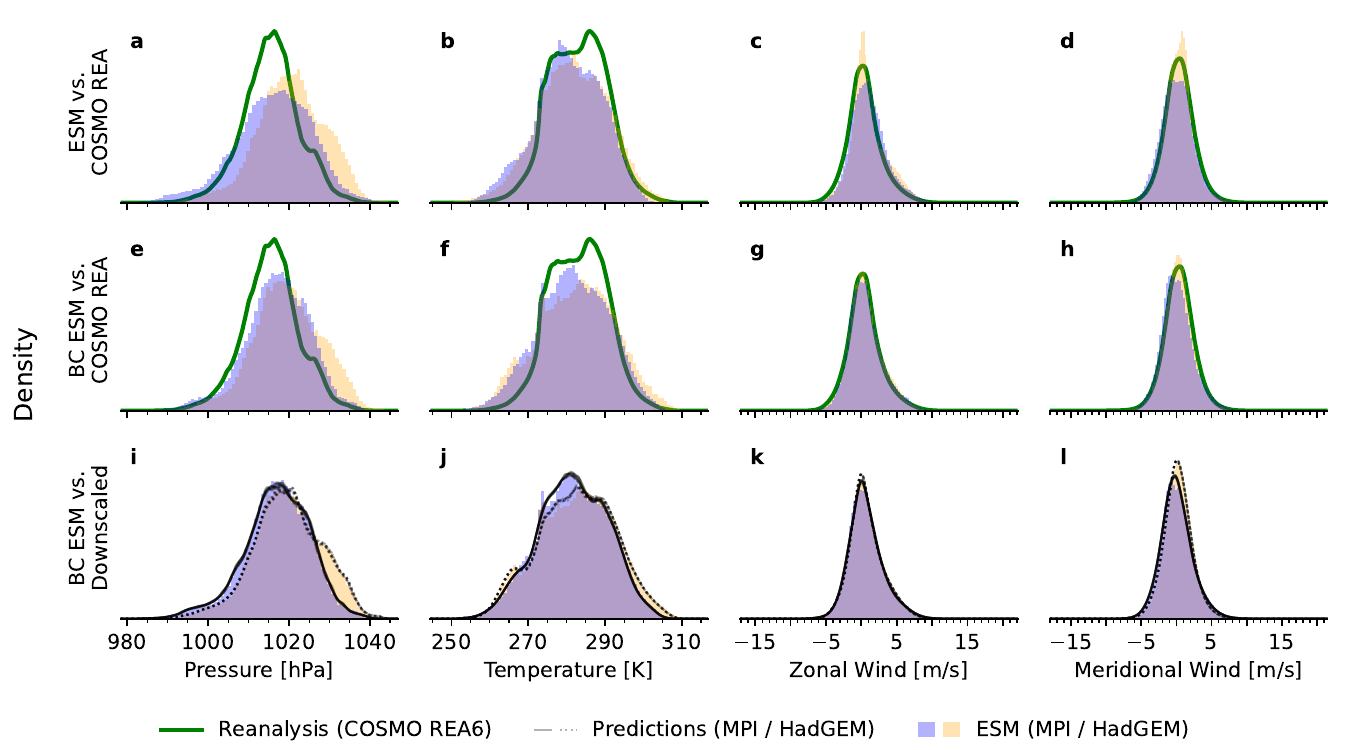}
  \caption{
    \textbf{Comparison of value distributions: ESM, de-biased ESM, reanalysis data, predictions.}
    This plot visualizes aggregated value distributions of reanalysis data (green), ESM simulations (two different models: MPI in purple, HadGEM in yellow), and downscaled predictions (black).
    The \emph{top row} compares the raw, uncorrected ESM simulations with the reanalysis data to visualize the biases in the respective climate distributions.
    The \emph{middle row} shows the bias-adjusted ESM simulations alongside the same reanalysis data to visualize the effect of the quantile-mapping procedure.
    The \emph{bottom row} compares the value distribution of the bias-adjusted ESM simulations (same as middle row) with their downscaled counterpart (black).
    The eight solid/dotted lines correspond to downscaled predictions of the MPI/HadGEM model.
    Two things are demonstrated: Comparing the \textit{top row} and the \textit{middle row} shows that the distribution mismatch between both ESM ensembles and the reanalysis data (\labelize{a}--\labelize{d}) is mitigated through the bias-correction step (\labelize{e}--\labelize{h}).
    There is some mismatch remaining, likely due to the short evaluation period.
    Secondly, the \textit{bottom row} shows that the distribution of the downscaled climate trajectories aligns with the coarse model input.
The considered time range is the year 2014. The model predicts 1-hourly steps starting January 01 at 06:00 AM and ending December 31 at 06:00 AM based on the corresponding 6-hourly coarse input. Visualizing two distinct CMIP6 ensembles (MPI and HadGEM) allows a comparison of the distributional mismatch a) between the respective climate outputs (purple vs. yellow) and b) between climate outputs and reanalysis data.
}
\label{fig:clim-downscaling-hist}
\end{figure}

\begin{figure}
\centering
\includegraphics[width=.49\linewidth]{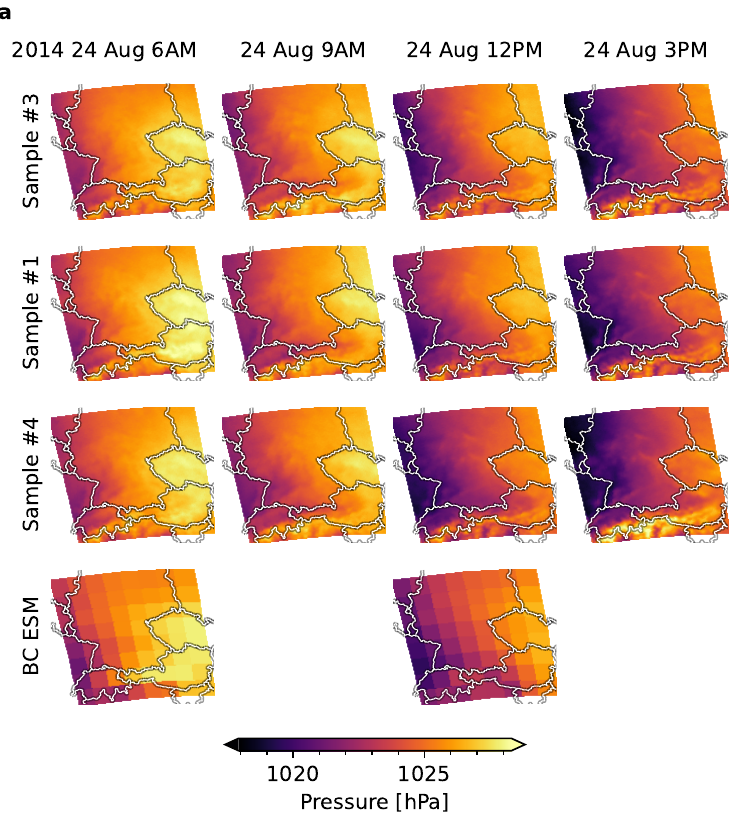}\hfill%
\includegraphics[width=.49\linewidth]{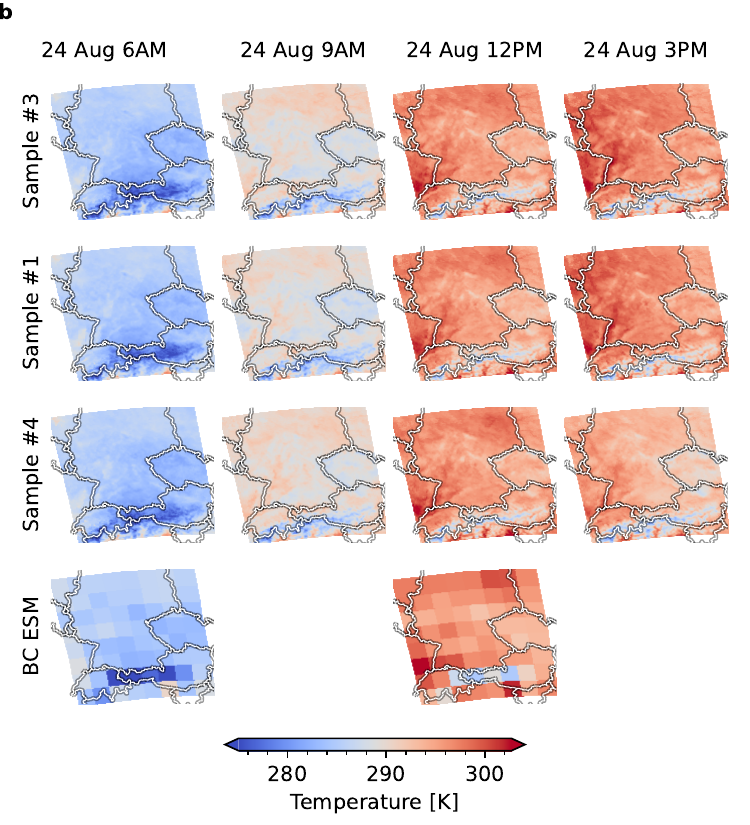}\\
\includegraphics[width=.49\linewidth]{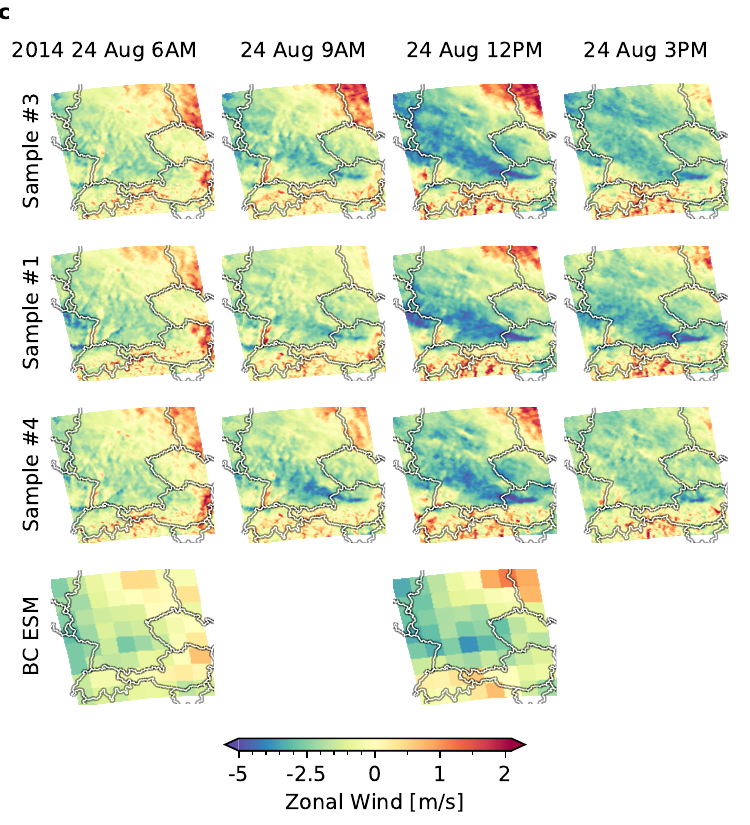}\hfill%
\includegraphics[width=.49\linewidth]{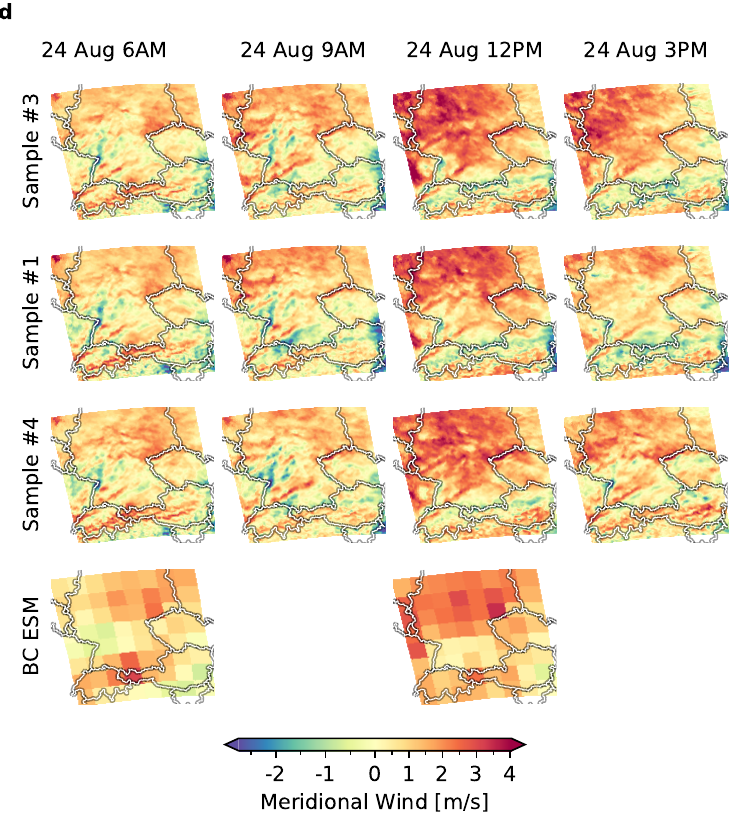}
\caption{
\textbf{Local climate information through downscaling.}
The plot shows spatiotemporally downscaled predictions for the MPI-HR ensemble that is part of the CMIP6 simulations.
Time progresses from left to right in three-hourly steps, starting on August 24, 2014, at 06:00 AM and ending the same day at 03:00 PM.
The four variables, mean sea-level pressure (\labelize{a}), surface temperature (\labelize{b}), zonal (\labelize{c}) and meridional (\labelize{d}) wind-speed components, are downscaled jointly by the model.
The top three rows show the progression of three randomly selected samples. The bottom row shows the corresponding conditioning information from the coarse bias-corrected (BC) ESM simulations.
Where the observation is blank, the model interpolates in time, without any conditioning information.
}
\label{fig:clim-downscaling-grid}
\end{figure}

The above experiments evaluated the model in a setting that ensures that the large-scale predictor perfectly matches the statistics of the fine-scale reanalysis data.
We proceed to downscaling climate-model outputs, using 6-hourly CMIP6 ESM simulations as conditioning information.
As in the previous experiments, the four variables considered are downscaled spatially (by a factor of $16\times 16$) and temporally (by a factor of $6$) to align with the resolution of the reanalysis data.
The same statistical relationship (\cref{eq:stat-relationship-obsmodel}) between coarse input and downscaled output is assumed in this experiment.
Downscaling climate-model outputs, however, presents a greater challenge because there is no direct pairing between coarse- and fine-scale climate data, which renders the comparison of our downscaling predictions to a ground truth impossible.
The goal of this experiment is to predict local patterns on a fine spatiotemporal grid, assuming two distinct perfect predictors given by two realizations from the distribution of coarse climate simulations.
We evaluate whether our downscaling model---while adhering to the statistical relationship imposed between each prediction and different coarse ESM simulations (\cref{fig:clim-downscaling-hist})---is capable of simultaneously predicting local climate (\cref{fig:clim-downscaling-grid}).
Notably, climate model biases are not encoded directly and therefore not addressed by the downscaling model.
As motivated by \citet{Volosciuk2017bcds}, downscaling can be separated into, firstly, mitigating climate model biases and, secondly, bridging the gap between coarse and fine grids.
We adopt this perspective, thereby focusing almost exclusively on the latter.
As a pre-processing step, we apply a per-variable quantile-mapping bias-correction \cite{maraun2013bias} to the ESM outputs.

\Cref{fig:clim-downscaling-hist} \labelize{a}--\labelize{d} shows two model ensembles of the four variables from different climate models: the MPI-HR (purple) and the HadGEM model (yellow) for the year 2014 and, in comparison, the reanalysis data distribution (green curve).
Comparing the distributions of the two ESMs, it becomes evident that, even in terms of their aggregated-value distribution, climate model outputs can differ significantly from each other and, moreover, each individual ESM differs substantially from the reanalysis data.
\Cref{fig:clim-downscaling-hist} \labelize{e}--\labelize{h} visualizes the effect of the quantile-mapping bias adjustment, which adjusts the marginal distributions of each variable in the ESM simulations to better match those of the reanalysis data.
As a result, the bias-corrected ESM simulations more closely align with the reanalysis data (green curve). A comparison of the uncorrected and bias-corrected distributions (\cref{fig:clim-downscaling-hist} \labelize{a}--\labelize{d} vs. \labelize{e}--\labelize{h}) reveals that some residual differences remain.
\Cref{fig:clim-downscaling-hist} \labelize{i}--\labelize{l} visualizes the distribution of eight downscaled predictions (black) for each of the two climate models.
The downscaled predictions for MPI-HR (solid lines) and HadGEM (dotted lines) closely match the value distributions of their respective bias-corrected ESM simulations.
Consequently, the downscaling model preserves the statistical properties of the coarse ESM input: the downscaled distributions are neither shifted nor skewed relative to their bias-corrected coarse counterparts, and the alignment extends to the tails of the distributions. As a result, any temporal changes or variations in the distribution shape present in the coarse input are faithfully retained in the downscaled output.

\Cref{fig:clim-downscaling-grid} presents a qualitative assessment of the spatial and temporal progression of downscaled ESM simulations.
The model output reproduces spatial patterns, which are, for example, consistent with geographical features, such as the Alps, and temporal patterns, like the day-night cycle, while introducing local weather dynamics consistent with the observations.
Together, \cref{fig:clim-downscaling-hist} and \cref{fig:clim-downscaling-grid} show that the model can generate high-resolution weather trajectories that introduce complex local dynamics while preserving the value distribution of the respective ESM simulations.


\section*{Discussion}

We introduced a probabilistic approach to joint spatial and temporal downscaling of multiple variables from climate to weather scale.
The presented framework revolves around a generative diffusion model, which is trained to learn an implicit representation of the dynamical patterns in reanalysis data and serves as a probabilistic emulator for the forward dynamics model.
By conditioning the forward model via an observation model on climate-model output, we can sample from a posterior downscaling distribution.
Each sample drawn from this posterior adheres to the established statistical relationship between climate output and fine-scale weather and avoids inconsistencies between time steps and variables, which would lead to physically implausible behavior.

Our approach aligns with a broader trend \cite{Adewoyin2021,Harder2023,Aich2024,Hess2025,Bischoff2024unpaireddownscaling,schmidt2024wind} of replacing computationally expensive simulations with statistical models that emulate earth-system dynamics.
\Citet{Rampal2024enhancing} give a topical overview and discussion regarding the use of ML methods for statistical downscaling.
In particular, generative-modeling techniques, such as normalizing flows \cite{groenke2020climalign}, GANs \cite{Hess2022physically,harris2022downscaling}, and diffusion models \cite{Aich2024,Hess2025,Tomasi2025} have been emerging as a popular model class.
Our model extends existing research \citep[for example][]{schmidt2024wind,langguth2024,Aich2024,Hess2025,addison2024machine} by enforcing coherence across spatial, temporal, and variable dimensions.
In particular, our predictions are sampled from a \emph{joint} distribution, avoiding the disconnection that typically comes with sampling sequences of temporally independent states for each individual atmospheric variable.
Inconsistencies between downscaled time steps and variables make predicted time series as a whole physically implausible and render the predictions unsuitable for downstream applications that require coherent estimates.
We present our framework as one possible technique for such coherence-dependent applications, which could include, e.g., driving dynamical climate impact models and studies of compound events \cite{Zscheischler2018}, which require inter-variable coherence.
A recent related method by \citet{srivastava2024precipitation} treats the spatial and temporal domain jointly while focusing on a single variable (precipitation).
For a chosen, fixed sequence length, their pipeline generates high-resolution estimates from coarse simulations by separating the task into, first, a deterministic statistical downscaling model, followed by a generative model that introduces probabilistic estimates of high-frequency patterns.
A related model class are conditional weather generators \cite{wilks1999weathergenerator,Rampal2024enhancing}\citep[Chapter 13]{maraun2018dsbook}, which require substantial expertise and resources to implement.
By training an unconditional generative model, we differ from previous approaches that integrate the conditioning directly into the training process \citep{Aich2024,schmidt2024wind, Tomasi2025}.
In contrast to \citet{Harder2023}, our method enables soft and uncertainty-aware constraints \emph{post-training} through posterior inference.
\Citet{Hess2025} exploit the iterative-denoising aspect of diffusion models and initialize the generative process at an only partially perturbed target state.
This allows conditioning the model on large-scale information, which the model enhances by replacing the remaining noise with local patterns.
Notably, this notion of "conditioning" lacks a clear probabilistic interpretation and is thus fundamentally different from how we use the term throughout this work.
Further, existing work focuses on mapping between different reanalysis data sets \citep{Harder2023, Winkler2024}, whereas we explicitly developed a model that allows mapping different climate scenarios to finely resolved time series that predict local weather patterns.
Our trained model can be readily re-used for mapping from different coarse predictors to the target resolution without re-training, simply by adapting the observation model and input data accordingly.
Diffusion bridges have been proposed for unpaired downscaling of fluid dynamics, though their application to climate-model downscaling remains to be demonstrated \cite{Bischoff2024unpaireddownscaling}.

The intended scope of our work is to demonstrate the score-based data assimilation framework by \citet{sda2023rozet} as an elegant and flexible technique for probabilistic downscaling and, perspectively, for other inference problems in the context of atmospheric dynamics.
More than outperforming existing per-variable and per-time-step downscaling methods, or establishing our downscaling framework as a new, universally favored approach, we aim to provide a framework for downstream applications that
\begin{enumerate}
  \item require spatio-temporally coherent time series,
  \item consider multiple variables jointly,
  \item benefit from flexibly encoding additional prior knowledge about the relation between input and output through the post-training conditioning.
\end{enumerate}
The comprehensive feature set offered by our model complicates direct comparisons with existing methods. Nevertheless, we demonstrate that, according to standard statistical distance metrics, our framework maintains competitive performance when evaluated per time step and per variable against benchmark approaches (see Supplementary Table 1). Further, spectral density comparisons between predictions and reanalysis data are presented in Supplementary Section 2, and a concise assessment of predictive skill for wind-power generation is provided in Supplementary Section 4.
Supplementary Figure 11 demonstrates that the model uses learned relationships between the jointly processed variables.
Looking ahead, we anticipate that the expanded capabilities of the presented framework will broaden its applicability to a wider range of use cases requiring coherent, high-resolution climate information.
Additionally, the modular structure of our approach (\cref{fig:pipeline-sketch}) enables the straightforward integration of further, potentially complex, domain-specific knowledge via the observation model in future work.
For example, established functional relationships between predicted climate variables and external forcings could be incorporated without retraining the generative model.
This modularity allows the trained score model to serve as a reusable foundation for subsequent studies. The diffusion model, including trained weights, and the implementations of the methods and experiments are provided at \url{https://github.com/schmidtjonathan/Climate2Weather}.

The main limitation of the presented framework is the increased computational cost that comes with coherent predictions.
While we have extended the original method by \citet{sda2023rozet} in terms of scalability, enabling it to process substantially longer trajectories, the temporal and spatial extent of the study is still limited by computational demands.
Due to the simultaneous processing of the entire state space, memory requirements remain a limiting factor.
This study limits itself to a small region (inspired by \citet{langguth2024}) with a diverse range of orography to establish and validate the methodological framework.
However, we believe that long-range teleconnections that affect the sub-region under study are, to some extent, reflected in the regional predictions, given that the training data has been drawn from a Europe-wide reanalysis dataset.
This argument is visually supported by embedding high-resolution predictions into a larger spatial context of reanalysis data in Supplementary Section 3 (Supplementary Figures 3 to 6).
We conclude that the technique is likely most useful for smaller study regions, for which highly accurate predictions of local dynamics are desired.

In summary, our framework enables joint spatial and temporal downscaling of multiple climate variables, producing coherent, high-resolution scenarios from coarse climate simulations. By decoupling the learning of dynamical patterns from the conditioning on new inputs, the model offers a flexible and efficient tool for inference tasks in meteorology and climate science. This approach facilitates the use of long-term climate projections for local impact studies across multiple timescales.



\section*{Methods}

\subsection*{Data}

We analyze four interrelated atmospheric variables, namely the zonal (\texttt{uas}) and meridional (\texttt{vas}) components of near-surface (10 meter) wind speeds, surface (2 meter) air temperature (\texttt{tas}), and sea level pressure (\texttt{psl}) due to their crucial importance for understanding atmospheric dynamics. They evolve on different spatial and temporal scales, enabling insights into the performance of downscaling methods across different scales.

For training (2006-2013) and evaluation (2014) of the model, we use data from the Consortium for Small-Scale Modeling Reanalysis 6 (COSMO-REA6) data set, a high-resolution reanalysis product for the European domain developed by the German Weather Service (Deutscher Wetterdienst; DWD) \cite{cosmo2015bollmeyer} with a spatial resolution of approximately $6~$km and hourly temporal resolution. It serves as the ground truth observational dataset in the perfect-predictor evaluation \cite{maraun2015VALUE} of the presented methodological framework.
The COSMO-REA6 data contains errors for some variables and lacks some observations for the years prior to 2006. In this work, only data from 2006 onwards was used.

We apply our downscaling model to historical model runs from two established general circulation models (GCMs) that are part of the sixth phase of the Coupled Model Intercomparison Project (CMIP6) \citep{Eyring2016}, namely the higher-resolution earth system model (ESM) of the Max--Planck Institute (MPI-ESM1.2-HR) \citep{muller2018higher} and the high-resolution configuration of the third Hadley Centre Global Environment Model in the global coupled configuration 3.1 (HadGEM3-GC3.1-LM) \citep{andrews2020historical}. Both models have an approximately $100~$km spatial and 6-hourly temporal resolution.

We restrict our analysis to a spatial subregion that is oriented at the benchmark region proposed by \citet{langguth2024}. The region includes parts of Germany, Switzerland, Austria, and the Czech Republic ($6\degree{E}-16\degree{E}$, $46\degree{N}-52\degree{N}$ (see Supplementary Figure 1) which results in $128\times 128$ grid points per time point for the reanalysis dataset.
  We use data for the period 2006-2014, the time range in which the reanalysis and the historical GCM runs overlap.
  The GCM data is spatially re-gridded to a rotated latitude-longitude grid using bilinear interpolation to match the coordinates and grid type of the reanalysis dataset.
  As a pre-processing step, we apply a quantile-mapping procedure to the GCM data.
  This mitigates biases by adjusting the value distribution of the GCM to align better with the distribution of the reanalysis product.
  For this, we use the quantile-mapping implementation provided by the \texttt{python-cmethods} \cite{pythoncmethods} Python package with default parameters.

  \subsection*{Model}

  We approach statistical downscaling as a Bayesian-inference problem \cite{gelman2013bayesian,bishop2006pattern}:
  An uncertain estimate for an unknown quantity (here: high-resolution weather) is obtained by first formulating a prior distribution that encodes known, assumed, or learned properties of the unknown quantity.
  The prior defines the output space of the prediction model.
  Through an observation model (or "likelihood"), the prior is conditioned on the available information (here: coarsely simulated climate output) to yield the posterior distribution.
  The posterior ideally captures the uncertainty that is both inherent in the prior and which arises from incomplete information and model mismatch.

  For inference in the context of dynamical systems, it is useful to formulate a prior model that somehow represents a mechanistic system that we assume to underlie the unknown dynamics.
  The core of the presented methodological framework is a score-based diffusion model (DM) \citep{dm2015sohldickstein,ddpm2020ho,sdedm2021song,variationaldms2021kingma,edm2022karras}, which constitutes the prior model of the framework.
  DMs are an instance of generative models that involve training a deep neural network on a finite set of data points, which can then be used to generate new samples from the underlying data distribution.
  The basic formulation of DMs involves two main components. Firstly, a forward \emph{diffusion process} transports the data distribution to a known, tractable distribution, such as a standard normal distribution \citep{ddpm2020ho,sdedm2021song}.
  In this process, structured data points (like RGB images, weather states, etc.) are successively perturbed until all signals are entirely replaced by noise.
  The pivotal insight is that a certain class of diffusion processes are known to have a reverse counterpart, providing a generative process that successively refines samples from the noise distribution into structured data.
  The reversal of the diffusion process requires access to the \emph{score function}, which can be understood as a function that, for any given degree of perturbation during the reverse diffusion process, separates noise from the signal to remove it.
  An exemplary such de-noising process for sequences of meridional wind speeds is visualized in Supplementary Figure 10.
  In the following, we will give a brief overview of the general framework and the extensions to it necessary to obtain the results presented in this work.

  The considered class of diffusion processes are Gauss--Markov processes that are the solution to linear stochastic differential equations of the form \citep{sdebook2019sarkka}
  \begin{equation}\label{eq:forward-diffusion-process}
    \dif X(\difftime) = \mat{F}X(\difftime)\dif \difftime + \mat{L}\dif W(\difftime), \quad X(0) \sim \normal{X_0}{\Sigma_0}.
  \end{equation}
  The state $X(\difftime)$ at $\tau=0$ is a data point $X_0 \sim \mathcal{D}$ selected from a data set $\mathcal{D}$. The data point is successively perturbed through the diffusion process and thus loses all structure as $\tau \rightarrow \mathrm{T}$.
  Note that $\difftime$ is sometimes referred to as "time", which does not mean physical time but rather a continuous degree of perturbation of the initial state.
  Steps in physical time will later be denoted in the subscript, e.g., $X_{1:L}(\difftime) := (X_1(\difftime), \dots, X_L(\difftime))$ will denote a time series of length $L$ perturbed according to $\difftime$.
  The drift $\mat{F}$ and dispersion $\mat{L}$ define functional properties of the forward process, which is driven by Brownian motion $W(\difftime)$. At a final time step $\mathrm{T}$ (often $\mathrm{T}=1$), the process converges to a Gaussian distribution such that $X(\mathrm{T}) \sim \mathcal{N}\left(0, \mat{\Pi}\right)$, where the final-step covariance $\mat{\Pi}$ depends on the choice of $\mat{F}$ and $\mat{L}$ and is often modeled to be the identity matrix.
  From a result by \citet{reversediff1982anderson}, the reverse process of \cref{eq:forward-diffusion-process} is known and given as
  \begin{equation}\label{eq:backward-diffusion-process}
    \dif X(\difftime) = \left[\mat{F}X(\difftime) - \mat{L}\mat{L}^\top \underbrace{\nabla_{X(\difftime)}\log p_\tau(X(\difftime))}_{\text{score function}} \right] \dif \difftime + \mat{L}\dif \overleftarrow{W}(\difftime), \quad X(\mathrm{T}) \sim \mathcal{N}\left(0, \mat{\Pi}\right).
  \end{equation}
  This process is driven by reverse Brownian motion $\overleftarrow{W}(\difftime)$ and depends on the gradient of the log-marginal density, called the \emph{score}, of the diffused state $X(\difftime)$ at every perturbation-time point $\difftime$.
  We call \cref{eq:backward-diffusion-process} the \emph{generative process} as it allows sampling unseen data points from the data distribution simply by
  \begin{enumerate}
    \item sampling a vector of independent Gaussian noise $X(\mathrm{T}) \sim \mathcal{N}\left(0, \mat{\Pi}\right)$
    \item numerically simulating \cref{eq:backward-diffusion-process} backwards through time, starting from $X(\mathrm{T})$ and ending at a generated sample $X(0)$.
  \end{enumerate}
  There is much existing and active research regarding the sampling algorithm used in step 2., which, in general, is slower and more complex when comparing diffusion models to other methods from the generative-model class \citep{zhang2023fast,consistency2023song,pmlr-v202-zheng23d}.
  \citet{Hess2025} employ a recent extension to the diffusion-model framework that allows single-step sampling.
  This work uses a standard technique that solves an ordinary differential equation related to the marginal distribution of \cref{eq:backward-diffusion-process} \citep{ddim2021song,sdedm2021song,edm2022karras} following the original proposal by \citet{sda2023rozet}.

  In practice, the score of $p_\tau(X(\difftime))$ is not accessible and has to be estimated. It is common practice to train a noise-dependent neural network $s_\theta(X(\difftime), \difftime)$ on a finite set of samples from the data distribution to approximate the true, intractable score function.
  Most practical approaches optimize a de-noising score-matching objective \citep{denoisingsm2011vincent,ddpm2020ho,ddim2021song,sdedm2021song,variationaldms2021kingma,edm2022karras}.
  In light of this, simulating \cref{eq:backward-diffusion-process} is commonly perceived as iteratively de-noising the initial Gaussian random state $X(\mathrm{T})$ and successively refining it to a noise-free data point.
  For spatial data, a time-conditioned U-Net \citep{unet2015ronneberger,ddpm2020ho} architecture has proven effective in modelling the score function.
  The U-Net architecture consists of multiple levels, each containing blocks of convolutional layers and skip connections. Along these levels, the spatial dimensionality of the input data is first successively reduced and then increased again.
  This encoder-decoder structure results in a bottleneck layer that forces the model to extract a limited set of meaningful features from the input.
  At each level, skip connections between equal-resolution blocks are introduced to facilitate the learning task by maintaining information throughout the encoding- and decoding process.
  This work uses a U-Net architecture with a total number of around $72$ million parameters, including a self-attention mechanism at the $8\times 8$-bottleneck layer \citep{ddpm2020ho,guidance2021dhariwal}.

  The framework used in the presented experiments leverages the work by \citet{sda2023rozet} that extends the basic DM framework, enabling robust sampling from a posterior distribution and generating arbitrary-length sequences of spatiotemporally coherent state trajectories.
  The idea behind score-based data assimilation (SDA) is to separate an inference task into two parts:
  \begin{enumerate}[label=\Roman*.]
    \item learning a generative model that generates spatiotemporal patterns from the target output space and
    \item conditioning the prior model using a functional or statistical constraint (the observation model) that defines the specific inference task.
  \end{enumerate}
  This separation aligns with the idea of Bayesian inference, in which a posterior estimate is obtained from combining a prior model (I.) with an observation model (II.) through conditioning.
  A high-level introduction of both parts follows below. For more details regarding the score-based data assimilation method, an extensive evaluation, and the corresponding code base, we refer to the original publication by \citet{sda2023rozet}.

  The generative model (I.) constitutes the prior in the present Bayesian-inference approach to statistical climate downscaling.
  As described above, the diffusion model requires a trained score model $s_\theta(X(\difftime), \difftime)$ to iteratively refine an initial random-noise sample into a generated data point (cf.~\cref{eq:backward-diffusion-process}).
  In our case, the generative model outputs uninformed sequences of high-resolution weather patterns.
  To generate sequential data, \citet{sda2023rozet} propose to learn a de-noising model on short fixed-length time windows.
  During training, the time window is flattened into the channel-dimension of the convolutional neural network that represents the score model $s_\theta$.
  This allows the model to learn correlations between different variables at different points in time.
  In the sampling routine, each point in time is de-noised by computing the score in the context of a surrounding temporal context window of size $w = 2k + 1$.
  \Citet{sda2023rozet} argue that many dynamical systems are (approximately) Markovian and in order to predict the current state, it is sufficient to regard the Markov blanket of order $k$ around the current time step, instead of the entire trajectory.
  This yields output dynamics that are temporally coherent without the need for training the network on long sequences, which would not only significantly increase the computational complexity but also fix the desired sequence length $L$ as a pre-determined hyperparameter, greatly limiting flexibility.
  Through an intricate convolution-like routine, the windows are assembled into trajectories of arbitrary length $L$.
  The corresponding pseudo-code is taken from \citet{sda2023rozet} and given in \cref{alg:composing}.

  \begin{algorithm}[H]
    \caption{Composing the sequential score function $s_\theta(X_{1:L}(\difftime), \difftime)$} \label{alg:composing}
    \begin{algorithmic}[1]
      \State \textbf{Input}: sample $X_{1:L}(\difftime)$ of sequence length $L$, perturbation level $\difftime$;\\ Markov order $k$; trained score model $s_\theta$.
      \State \textbf{Output}: Score $s_{1:L} \approx \nabla_{X_{1:L}(\difftime)} \log p_\difftime(X_{1:L}(\difftime))$ for every time step $t \in 1{:}L := (1, \dots, L)$
      \Function{$s_\theta(X_{1:L}(\difftime), \difftime)$}{}
      \State $s_{1:k+1} \gets s_\theta(X_{1:2k+1}(\difftime), \difftime)[:\!k + 1]$
      \For{$t = k + 2$ to $L - k - 1$}
      \State $s_t \gets s_\theta(X_{t-k:t+k}(\difftime), \difftime)[k+1]$
      \EndFor{}
      \State $s_{L-k:L} \gets s_\theta(X_{L-2k:L}(\difftime), \difftime)[k+1\!:]$
      \State \Return{$s_{1:L}$}
      \EndFunction{}
    \end{algorithmic}
  \end{algorithm}

  For our experiments, we require scaling the original SDA framework to allow for the prediction of very long time horizons with thousands of steps.
  Unfortunately, sampling a spatiotemporally coherent trajectory in the described manner comes at an increased computational cost and, in particular, significant memory demands since the entire state trajectory must be held in memory.
  To make the approach scale to predicting 1-hourly weather dynamics for multiple years, we implement the spatiotemporal score function \cref{alg:composing} as a massively parallelized convolution operation through time that efficiently manages the available memory---optionally on multiple devices.
  This extension of the original implementation has proven indispensable for the problem scale.
  The corresponding Python implementation is provided at \url{https://github.com/schmidtjonathan/Climate2Weather}.

  The generative model described above can generate random sequences of weather-like patterns.
  One can think of it as a "physical prior" in that it represents a statistical model for the general spatial and temporal patterns observed in high-resolution weather dynamics.
  Note that the physical laws are only implicitly represented by this model, through learning it from existing simulations, and no explicit mechanistic laws are encoded in the model.
  To tackle the downscaling task specifically, we have to inform the prior model about the climate constraints.
  To this end, we introduce the conditioning mechanism, which allows us to impose the constraints provided by the coarse ESM simulations onto the prior model.

  We proceed to describe the conditioning mechanism (II.).
  While the unconditioned diffusion model generates samples $p(X(0))$, the goal is to sample from a posterior $p(X(0) \mid Y)$, instead.
  We denote the high-resolution output with $X(0)$ and the coarse input with $Y$.
  Here, we include the noise level $\difftime$ of the sample $X(\difftime)$ into the notation (in the parentheses), as it will become relevant in this section.
  Recall that $X(0)$ is the generated, noise-free sample.
  An observation model $p(Y \mid X(0))$ relates the coarse input $Y$ to $X(0)$.
  We consider a Gaussian observation model
  \begin{equation}\label{eq:observation-model}
    p(Y \mid X(0)) = \mathcal{N}(Y; h(X(0)), \mat{R}),
  \end{equation}
  with an observation operator $h(X(0))$ that can be a linear or non-linear function \citep{dps2023chung,sda2023rozet} and observation- or sensor noise defined by a positive-definite matrix $\mat{R}$. In our case, $h$ selects from the high-resolution state $X$ every six hours and computes an area averaging in space (see \cref{eq:stat-relationship-obsmodel}).
  For our experiments, the observation noise $\mat{R}$ is selected by running a simple random search on a range of plausible values.
  To accelerate model selection, we chose $\mat{R}$ based on the model's downscaling performance on a two-day window.
  The resulting observation model is used in all presented experiments.
  Sampling from $p(X(0) \mid Y)$ amounts to replacing the score function in \cref{eq:backward-diffusion-process} by a posterior score function $\nabla_{X(\tau)}\log p_\tau(X(\tau) \mid Y)$.
  The posterior score is obtained through Bayes' rule by noting that
  \begin{align}
    \nabla_{X(\difftime)} \log p(X(\difftime) \mid Y) = \nabla_{X(\difftime)} \log p_\difftime(X(\difftime)) + \nabla_{X(\difftime)} \log p(Y \mid X(\difftime)),
  \end{align}
  The normalization constant is not dependent on $X$ and thus vanishes when taking the gradient.
  The prior score is modeled with the parametric score model $s_\theta(X(\difftime), \difftime)$.
  Note that the likelihood term $p(Y \mid X(\difftime))$ is defined on the perturbed state $X(\difftime)$ and has to be approximated to be compatible with the observation model $p(Y \mid X(0))$.
  For details, we refer the reader to \citep[Section 3.1]{dps2023chung}, who establish the concept of "Diffusion Posterior Sampling" and to its extension by \citet{sda2023rozet} (Section 3.2).
  Further, a mathematical description is given in Supplementary Section 6.
  To make the conditioning work efficiently at the problem scales considered in our experiments, we introduce an approximation in the conditioning mechanism that avoids the computation of gradients with respect to the score network. This is described and derived in Supplementary Section 6.

  \subsection*{Experimental Setup}\label{subsec:methods:exp-setup}

  ESM projections are not directly paired to reanalysis data, so we begin by evaluating the generative downscaling model in an on-model setting.
  We set up an experimental setting that allows us to compare the model output to a ground truth to assess its predictive performance.
  Firstly, the model is trained on a subset (2006--2013) of the COSMO reanalysis data.
  A separate subset (2014) serves as a test set. Then, \dots
  \begin{enumerate}
    \item \dots from the test set, we select an evaluation period and generate artificial coarse observations that represent ESM projections.
      The spatial resolution is reduced by a factor of $16\times 16$ and the temporal resolution by a factor of 6 (see \cref{eq:stat-relationship-obsmodel}). Specifically,
      \begin{enumerate}
        \item every 6th time step (hour) is selected from the ground truth,
        \item a patch-wise spatial averaging operation is applied throughout the spatial region at every time step. We compute the arithmetic mean for each $16\times 16$ spatial patch to yield a single spatial observation.
      \end{enumerate}
    \item With the observations from 1., the model predicts the underlying 1-hourly, high-resolution weather dynamics lost by coarsening the data. Multiple samples are drawn.
    \item The samples are compared to the reanalysis data from the same time period and evaluated. Variations between the samples provide structured uncertainty.
  \end{enumerate}

  As a second step, ESM (CMIP 6) simulations replace the artificially coarsened reanalysis data for the final experiment.
  Instead of artificially spatiotemporally subsampled reanalysis data, we condition the score model on two different ensembles from the CMIP 6 data set: the MPI-HR and the HadGEM3 runs.
  The experimental setup is adopted exactly from the on-model experiments, aside from the ESM conditioning information and the extended considered time horizon of one full year.
  Concretely, we downscale the ESM simulations for the considered spatial patch from January 1, 06:00 AM until December 31, 06:00 AM ($8\,736$hours) for the year 2014, increasing the spatial resolution from $8\times 8$ ($\sim 100$km) to $128 \times 128$ ($\sim 6$km) grid points.
  For the four high-resolution variables, this corresponds to a total of $4 \times 8\,736 \times (128 \times 128) = 572\,522\,496$ predicted values given conditioning information that is coarser by a factor of $(16 \times 16) \times 6 = 1\,536$.
  For this specific experiment, we used four NVIDIA A-100 GPUs, each of which generated two predictions in parallel. Generating one 1-year sample of hourly downscaled predictions for the $128\times 128$-node region takes around two hours on a single NVIDIA A-100 device.


\section*{Data Availability}
The reanalysis data were taken from the COSMO REA-6 reanalysis product \citep{cosmo2015bollmeyer}.
The CMIP6 data \citep{Eyring2016} was downloaded from the Earth System Grid Federation (ESGF) at \url{https://esgf-node.llnl.gov/projects/cmip6/}.

\section*{Code availability}
The code (in Python) for model training and evaluation, data processing, and for reproducing all presented experiments and figures contained in this manuscript, is available at \url{https://github.com/schmidtjonathan/Climate2Weather}.
For the qunatile-mapping bias adjustment, the Python library \texttt{python-cmethods} \cite{pythoncmethods} was used.
Part of the plots were generated using the \texttt{cartopy} library \citep{Cartopy}.
The neural-network model and training are implemented using the \texttt{PyTorch} library \cite{pytorch}.

\section*{Acknowledgements}
The authors gratefully acknowledge financial support by the European Research Council through ERC CoG Action 101123955 ANUBIS; the DFG Cluster of Excellence "Machine Learning - New Perspectives for Science", EXC 2064/1, project number 390727645; the German Federal Ministry of Education and Research (BMBF) through the Tübingen AI Center (FKZ: 01IS18039A); the DFG SPP 2298 (Project HE 7114/5-1), and the Carl Zeiss Foundation, (project "Certification and Foundations of Safe Machine Learning Systems in Healthcare"), as well as funds from the Ministry of Science, Research and Arts of the State of Baden-Württemberg. The authors thank the International Max Planck Research
School for Intelligent Systems (IMPRS-IS) for supporting Jonathan Schmidt and Luca Schmidt.

\section*{Author contribution statement}
The joint project was initiated by J.S. and N.L. and coordinated by P.H. and N.L.
J.S. implemented the code base for the  model, training, experiments, data-processing and -pipeline (reanalysis data), as well as for evaluation and visualization of the results. J.S., L.S., and F.S. designed the experiments. L.S. implemented the data-processing for the climate simulations. The first version of the article was written by J.S. and L.S., after which all authors reviewed and edited the manuscript.

\section*{Competing Interests}
The authors declare that they have no competing interests.

\renewcommand{\bibname}{Reference}
\bibliography{bib}

\begin{thebibliography}{82}
\providecommand{\natexlab}[1]{#1}
\providecommand{\url}[1]{\texttt{#1}}
\expandafter\ifx\csname urlstyle\endcsname\relax
  \providecommand{\doi}[1]{doi: #1}\else
  \providecommand{\doi}{doi: \begingroup \urlstyle{rm}\Url}\fi

\bibitem[Bauer et~al.(2015)Bauer, Thorpe, and Brunet]{Bauer2015}
Peter Bauer, Alan Thorpe, and Gilbert Brunet.
\newblock {The quiet revolution of numerical weather prediction}.
\newblock \emph{Nature}, 525:\penalty0 47--55, September 2015.
\newblock ISSN 1476-4687.

\bibitem[Moon et~al.(2024)Moon, Streffing, Lee, Semmler, Andr\'es-Mart\'{\i}nez, Chen, Cho, Chu, Franzke, G\"artner, Ghosh, Hegewald, Hong, Koldunov, Lee, Lin, Liu, Loza, Park, Roh, Sein, Sharma, Sidorenko, Son, Stuecker, Wang, Yi, Zapponini, Jung, and Timmermann]{moon2024globalhresm}
J.-Y. Moon, J.~Streffing, S.-S. Lee, T.~Semmler, M.~Andr\'es-Mart\'{\i}nez, J.~Chen, E.-B. Cho, J.-E. Chu, C.~Franzke, J.~P. G\"artner, R.~Ghosh, J.~Hegewald, S.~Hong, N.~Koldunov, J.-Y. Lee, Z.~Lin, C.~Liu, S.~Loza, W.~Park, W.~Roh, D.~V. Sein, S.~Sharma, D.~Sidorenko, J.-H. Son, M.~F. Stuecker, Q.~Wang, G.~Yi, M.~Zapponini, T.~Jung, and A.~Timmermann.
\newblock Earth’s future climate and its variability simulated at 9 km global resolution.
\newblock \emph{EGUsphere}, 2024:\penalty0 1--46, 2024.
\newblock \doi{10.5194/egusphere-2024-2491}.
\newblock URL \url{https://egusphere.copernicus.org/preprints/2024/egusphere-2024-2491/}.

\bibitem[Eyring et~al.(2016)Eyring, Bony, Meehl, Senior, Stevens, Stouffer, and Taylor]{Eyring2016}
Veronika Eyring, Sandrine Bony, Gerald~A. Meehl, Catherine~A. Senior, Bjorn Stevens, Ronald~J. Stouffer, and Karl~E. Taylor.
\newblock {Overview of the Coupled Model Intercomparison Project Phase 6 (CMIP6) experimental design and organization}.
\newblock \emph{Geosci. Model Dev.}, 9\penalty0 (5):\penalty0 1937--1958, May 2016.
\newblock ISSN 1991-959X.

\bibitem[Maraun and Widmann(2018)]{maraun2018dsbook}
Douglas Maraun and Martin Widmann.
\newblock \emph{Statistical Downscaling and Bias Correction for Climate Research}.
\newblock Cambridge University Press, 2018.

\bibitem[Laprise(2008)]{Laprise2008}
Ren{\ifmmode\acute{e}\else\'{e}\fi} Laprise.
\newblock {Regional climate modelling}.
\newblock \emph{J. Comput. Phys.}, 227\penalty0 (7):\penalty0 3641--3666, March 2008.
\newblock ISSN 0021-9991.
\newblock \doi{10.1016/j.jcp.2006.10.024}.

\bibitem[Tapiador et~al.(2020)Tapiador, Navarro, Moreno, S{\ifmmode\acute{a}\else\'{a}\fi}nchez, and Garc{\ifmmode\acute{\imath}\else\'{\i}\fi}a-Ortega]{Tapiador2020}
Francisco~J. Tapiador, Andr{\ifmmode\acute{e}\else\'{e}\fi}s Navarro, Ra{\ifmmode\acute{u}\else\'{u}\fi}l Moreno, Jos{\ifmmode\acute{e}\else\'{e}\fi}~Luis S{\ifmmode\acute{a}\else\'{a}\fi}nchez, and Eduardo Garc{\ifmmode\acute{\imath}\else\'{\i}\fi}a-Ortega.
\newblock {Regional climate models: 30 years of dynamical downscaling}.
\newblock \emph{Atmos. Res.}, 235:\penalty0 104785, May 2020.
\newblock ISSN 0169-8095.
\newblock \doi{10.1016/j.atmosres.2019.104785}.

\bibitem[Risser et~al.(2024)Risser, Rahimi, Goldenson, Hall, Lebo, and Feldman]{Risser2024}
Mark~D. Risser, Stefan Rahimi, Naomi Goldenson, Alex Hall, Zachary~J. Lebo, and Daniel~R. Feldman.
\newblock {Is Bias Correction in Dynamical Downscaling Defensible?}
\newblock \emph{Geophys. Res. Lett.}, 51\penalty0 (10):\penalty0 e2023GL105979, May 2024.
\newblock ISSN 0094-8276.
\newblock \doi{10.1029/2023GL105979}.

\bibitem[Wilks(1999)]{wilks1999weathergenerator}
D.~S. Wilks.
\newblock Multisite downscaling of daily precipitation with a stochastic weather generator.
\newblock \emph{Climate Research}, 11\penalty0 (2):\penalty0 125--136, 1999.

\bibitem[Ba\~no Medina et~al.(2020)Ba\~no Medina, Manzanas, and Guti\'errez]{gmd-13-2109-2020}
J.~Ba\~no Medina, R.~Manzanas, and J.~M. Guti\'errez.
\newblock Configuration and intercomparison of deep learning neural models for statistical downscaling.
\newblock \emph{Geoscientific Model Development}, 13\penalty0 (4):\penalty0 2109--2124, 2020.
\newblock \doi{10.5194/gmd-13-2109-2020}.
\newblock URL \url{https://gmd.copernicus.org/articles/13/2109/2020/}.

\bibitem[Baño-Medina et~al.(2021)Baño-Medina, Manzanas, and Gutiérrez]{BanoMedina2021}
Jorge Baño-Medina, Rodrigo Manzanas, and José~Manuel Gutiérrez.
\newblock On the suitability of deep convolutional neural networks for continental-wide downscaling of climate change projections.
\newblock \emph{Climate Dynamics}, 57\penalty0 (11):\penalty0 2941--2951, 2021.
\newblock ISSN 1432-0894.
\newblock \doi{10.1007/s00382-021-05847-0}.
\newblock URL \url{https://doi.org/10.1007/s00382-021-05847-0}.

\bibitem[Balmaceda-Huarte et~al.(2024)Balmaceda-Huarte, Baño-Medina, Olmo, and Bettolli]{Balmaceda-Huarte2024}
Rocío Balmaceda-Huarte, Jorge Baño-Medina, Matias~Ezequiel Olmo, and Maria~Laura Bettolli.
\newblock On the use of convolutional neural networks for downscaling daily temperatures over southern south america in a climate change scenario.
\newblock \emph{Climate Dynamics}, 62\penalty0 (1):\penalty0 383--397, 2024.
\newblock ISSN 1432-0894.
\newblock \doi{10.1007/s00382-023-06912-6}.
\newblock URL \url{https://doi.org/10.1007/s00382-023-06912-6}.

\bibitem[Babaousmail et~al.(2021)Babaousmail, Hou, Gnitou, and Ayugi]{BABAOUSMAIL2021105614}
Hassen Babaousmail, Rongtao Hou, Gnim~Tchalim Gnitou, and Brian Ayugi.
\newblock Novel statistical downscaling emulator for precipitation projections using deep convolutional autoencoder over northern africa.
\newblock \emph{Journal of Atmospheric and Solar-Terrestrial Physics}, 218:\penalty0 105614, 2021.
\newblock ISSN 1364-6826.
\newblock \doi{https://doi.org/10.1016/j.jastp.2021.105614}.
\newblock URL \url{https://www.sciencedirect.com/science/article/pii/S1364682621000754}.

\bibitem[Smith et~al.(2019)Smith, Eade, Scaife, Caron, Danabasoglu, DelSole, Delworth, Doblas-Reyes, Dunstone, Hermanson, et~al.]{smith2019robust}
DM~Smith, R~Eade, AA~Scaife, L-P Caron, G~Danabasoglu, TM~DelSole, T~Delworth, FJ~Doblas-Reyes, NJ~Dunstone, L~Hermanson, et~al.
\newblock Robust skill of decadal climate predictions.
\newblock \emph{Npj Climate and Atmospheric Science}, 2\penalty0 (1):\penalty0 13, 2019.

\bibitem[Doblas-Reyes et~al.(2021)Doblas-Reyes, Sörensson, Almazroui, Dosio, Gutowski, Haarsma, Hamdi, Hewitson, Kwon, Lamptey, Maraun, Stephenson, Takayabu, Terray, Turner, and Zuo]{doblasreyes2021linking}
F.J. Doblas-Reyes, A.A. Sörensson, M.~Almazroui, A.~Dosio, W.J. Gutowski, R.~Haarsma, R.~Hamdi, B.~Hewitson, W.-T. Kwon, B.L. Lamptey, D.~Maraun, T.S. Stephenson, I.~Takayabu, L.~Terray, A.~Turner, and Z.~Zuo.
\newblock Linking global to regional climate change.
\newblock In V.~Masson-Delmotte, P.~Zhai, A.~Pirani, S.L. Connors, C.~Péan, S.~Berger, N.~Caud, Y.~Chen, L.~Goldfarb, M.I. Gomis, M.~Huang, K.~Leitzell, E.~Lonnoy, J.B.R. Matthews, T.K. Maycock, T.~Waterfield, O.~Yelekçi, R.~Yu, and B.~Zhou, editors, \emph{Climate Change 2021: The Physical Science Basis. Contribution of Working Group I to the Sixth Assessment Report of the Intergovernmental Panel on Climate Change}, pages 1363--1512. Cambridge University Press, Cambridge, United Kingdom and New York, NY, USA, 2021.
\newblock \doi{10.1017/9781009157896.012}.

\bibitem[Morelli et~al.(2025)Morelli, Effenberger, Schmidt, and Ludwig]{morelli2024climate}
Sofia Morelli, Nina Effenberger, Luca Schmidt, and Nicole Ludwig.
\newblock {Climate data selection for multi-decadal wind power forecasts}.
\newblock \emph{Environ. Res. Lett.}, 20\penalty0 (4):\penalty0 044032, March 2025.
\newblock ISSN 1748-9326.
\newblock \doi{10.1088/1748-9326/adc01f}.

\bibitem[Chen et~al.(2021)Chen, Rojas, Samset, Cobb, Diongue~Niang, Edwards, Emori, Faria, Hawkins, Hope, Huybrechts, Meinshausen, Mustafa, Plattner, and Tréguier]{chen2021framing}
D.~Chen, M.~Rojas, B.H. Samset, K.~Cobb, A.~Diongue~Niang, P.~Edwards, S.~Emori, S.H. Faria, E.~Hawkins, P.~Hope, P.~Huybrechts, M.~Meinshausen, S.K. Mustafa, G.-K. Plattner, and A.-M. Tréguier.
\newblock Framing, context, and methods.
\newblock In V.~Masson-Delmotte, P.~Zhai, A.~Pirani, S.L. Connors, C.~Péan, S.~Berger, N.~Caud, Y.~Chen, L.~Goldfarb, M.I. Gomis, M.~Huang, K.~Leitzell, E.~Lonnoy, J.B.R. Matthews, T.K. Maycock, T.~Waterfield, O.~Yelekçi, R.~Yu, and B.~Zhou, editors, \emph{Climate Change 2021: The Physical Science Basis. Contribution of Working Group I to the Sixth Assessment Report of the Intergovernmental Panel on Climate Change}, pages 147--286. Cambridge University Press, Cambridge, United Kingdom and New York, NY, USA, 2021.
\newblock \doi{10.1017/9781009157896.003}.

\bibitem[Jain et~al.(2023)Jain, Scaife, Shepherd, Deser, Dunstone, Schmidt, Trenberth, and Turkington]{jain2023importance}
Shipra Jain, Adam~A Scaife, Theodore~G Shepherd, Clara Deser, Nick Dunstone, Gavin~A Schmidt, Kevin~E Trenberth, and Thea Turkington.
\newblock Importance of internal variability for climate model assessment.
\newblock \emph{npj Climate and Atmospheric Science}, 6\penalty0 (1):\penalty0 68, 2023.

\bibitem[Hess et~al.(2022)Hess, Drüke, Petri, Strnad, and Boers]{Hess2022physically}
Philipp Hess, Markus Drüke, Stefan Petri, Felix~M. Strnad, and Niklas Boers.
\newblock Physically constrained generative adversarial networks for improving precipitation fields from earth system models.
\newblock \emph{Nature Machine Intelligence}, 4\penalty0 (10):\penalty0 828--839, 2022.
\newblock ISSN 2522-5839.
\newblock \doi{10.1038/s42256-022-00540-1}.
\newblock URL \url{https://doi.org/10.1038/s42256-022-00540-1}.

\bibitem[Aich et~al.(2024)Aich, Hess, Pan, Bathiany, Huang, and Boers]{Aich2024}
Michael Aich, Philipp Hess, Baoxiang Pan, Sebastian Bathiany, Yu~Huang, and Niklas Boers.
\newblock {Conditional diffusion models for downscaling {\&} bias correction of Earth system model precipitation}.
\newblock \emph{arXiv}, April 2024.

\bibitem[Ho et~al.(2020)Ho, Jain, and Abbeel]{ddpm2020ho}
Jonathan Ho, Ajay Jain, and Pieter Abbeel.
\newblock Denoising diffusion probabilistic models.
\newblock In H.~Larochelle, M.~Ranzato, R.~Hadsell, M.F. Balcan, and H.~Lin, editors, \emph{Advances in Neural Information Processing Systems}, volume~33, pages 6840--6851. Curran Associates, Inc., 2020.

\bibitem[Song et~al.(2021{\natexlab{a}})Song, Sohl-Dickstein, Kingma, Kumar, Ermon, and Poole]{sdedm2021song}
Yang Song, Jascha Sohl-Dickstein, Diederik~P Kingma, Abhishek Kumar, Stefano Ermon, and Ben Poole.
\newblock Score-based generative modeling through stochastic differential equations.
\newblock In \emph{International Conference on Learning Representations}, 2021{\natexlab{a}}.

\bibitem[Song et~al.(2021{\natexlab{b}})Song, Meng, and Ermon]{ddim2021song}
Jiaming Song, Chenlin Meng, and Stefano Ermon.
\newblock Denoising diffusion implicit models.
\newblock In \emph{International Conference on Learning Representations}, 2021{\natexlab{b}}.

\bibitem[Rozet and Louppe(2023{\natexlab{a}})]{sda2023rozet}
Fran\c{c}ois Rozet and Gilles Louppe.
\newblock Score-based data assimilation.
\newblock In A.~Oh, T.~Naumann, A.~Globerson, K.~Saenko, M.~Hardt, and S.~Levine, editors, \emph{Advances in Neural Information Processing Systems}, volume~36, pages 40521--40541. Curran Associates, Inc., 2023{\natexlab{a}}.

\bibitem[Kingma and Welling(2014)]{vae2014kingma}
Diederik~P. Kingma and Max Welling.
\newblock Auto-encoding variational bayes.
\newblock In Yoshua Bengio and Yann LeCun, editors, \emph{2nd International Conference on Learning Representations, {ICLR} 2014, Banff, AB, Canada, April 14-16, 2014, Conference Track Proceedings}, 2014.

\bibitem[Goodfellow et~al.(2014)Goodfellow, Pouget-Abadie, Mirza, Xu, Warde-Farley, Ozair, Courville, and Bengio]{gan2014goodfellow}
Ian Goodfellow, Jean Pouget-Abadie, Mehdi Mirza, Bing Xu, David Warde-Farley, Sherjil Ozair, Aaron Courville, and Yoshua Bengio.
\newblock Generative adversarial nets.
\newblock In Z.~Ghahramani, M.~Welling, C.~Cortes, N.~Lawrence, and K.Q. Weinberger, editors, \emph{Advances in Neural Information Processing Systems}, volume~27. Curran Associates, Inc., 2014.

\bibitem[Dinh et~al.(2014)Dinh, Krueger, and Bengio]{Dinh2014}
Laurent Dinh, David Krueger, and Yoshua Bengio.
\newblock {NICE: Non-linear Independent Components Estimation}.
\newblock \emph{arXiv}, October 2014.
\newblock \doi{10.48550/arXiv.1410.8516}.

\bibitem[Dhariwal and Nichol(2021)]{guidance2021dhariwal}
Prafulla Dhariwal and Alexander~Quinn Nichol.
\newblock Diffusion models beat {GAN}s on image synthesis.
\newblock In A.~Beygelzimer, Y.~Dauphin, P.~Liang, and J.~Wortman Vaughan, editors, \emph{Advances in Neural Information Processing Systems}, 2021.

\bibitem[Cao et~al.(2022)Cao, Tan, Gao, Xu, Chen, Heng, and Li]{Cao2022}
Hanqun Cao, Cheng Tan, Zhangyang Gao, Yilun Xu, Guangyong Chen, Pheng-Ann Heng, and Stan~Z. Li.
\newblock {A Survey on Generative Diffusion Model}.
\newblock \emph{arXiv}, September 2022.
\newblock \doi{10.48550/arXiv.2209.02646}.

\bibitem[Koo and Kim(2023)]{Koo2023}
Heejoon Koo and To~Eun Kim.
\newblock {A Comprehensive Survey on Generative Diffusion Models for Structured Data}.
\newblock \emph{arXiv}, June 2023.
\newblock \doi{10.48550/arXiv.2306.04139}.

\bibitem[Rozet and Louppe(2023{\natexlab{b}})]{sdaqg2023rozet}
François Rozet and Gilles Louppe.
\newblock Score-based data assimilation for a two-layer quasi-geostrophic model.
\newblock \emph{ArXiv}, abs/2310.01853, 2023{\natexlab{b}}.

\bibitem[Klein et~al.(1959)Klein, Lewis, and Enger]{klein1959perfectprog}
William~H. Klein, Billy~M. Lewis, and Isadore Enger.
\newblock Objective prediction of five-day mean temperatures during winter.
\newblock \emph{Journal of Atmospheric Sciences}, 16\penalty0 (6):\penalty0 672 -- 682, 1959.
\newblock \doi{10.1175/1520-0469(1959)016<0672:OPOFDM>2.0.CO;2}.
\newblock URL \url{https://journals.ametsoc.org/view/journals/atsc/16/6/1520-0469_1959_016_0672_opofdm_2_0_co_2.xml}.

\bibitem[Maraun et~al.(2010)Maraun, Wetterhall, Ireson, Chandler, Kendon, Widmann, Brienen, Rust, Sauter, Themeßl, Venema, Chun, Goodess, Jones, Onof, Vrac, and Thiele-Eich]{maraun2010perfectprog}
D.~Maraun, F.~Wetterhall, A.~M. Ireson, R.~E. Chandler, E.~J. Kendon, M.~Widmann, S.~Brienen, H.~W. Rust, T.~Sauter, M.~Themeßl, V.~K.~C. Venema, K.~P. Chun, C.~M. Goodess, R.~G. Jones, C.~Onof, M.~Vrac, and I.~Thiele-Eich.
\newblock Precipitation downscaling under climate change: Recent developments to bridge the gap between dynamical models and the end user.
\newblock \emph{Reviews of Geophysics}, 48\penalty0 (3), 2010.
\newblock \doi{https://doi.org/10.1029/2009RG000314}.
\newblock URL \url{https://agupubs.onlinelibrary.wiley.com/doi/abs/10.1029/2009RG000314}.

\bibitem[Lessig et~al.(2023)Lessig, Luise, Gong, Langguth, Stadtler, and Schultz]{Lessig2023}
Christian Lessig, Ilaria Luise, Bing Gong, Michael Langguth, Scarlet Stadtler, and Martin Schultz.
\newblock {AtmoRep: A stochastic model of atmosphere dynamics using large scale representation learning}.
\newblock \emph{arXiv}, August 2023.
\newblock \doi{10.48550/arXiv.2308.13280}.

\bibitem[Lam et~al.(2023)Lam, Sanchez-Gonzalez, Willson, Wirnsberger, Fortunato, Alet, Ravuri, Ewalds, Eaton-Rosen, Hu, Merose, Hoyer, Holland, Vinyals, Stott, Pritzel, Mohamed, and Battaglia]{Lam2023}
Remi Lam, Alvaro Sanchez-Gonzalez, Matthew Willson, Peter Wirnsberger, Meire Fortunato, Ferran Alet, Suman Ravuri, Timo Ewalds, Zach Eaton-Rosen, Weihua Hu, Alexander Merose, Stephan Hoyer, George Holland, Oriol Vinyals, Jacklynn Stott, Alexander Pritzel, Shakir Mohamed, and Peter Battaglia.
\newblock {Learning skillful medium-range global weather forecasting}.
\newblock \emph{Science}, 382\penalty0 (6677):\penalty0 1416--1421, November 2023.
\newblock ISSN 0036-8075.
\newblock \doi{10.1126/science.adi2336}.

\bibitem[Price et~al.(2025)Price, Sanchez-Gonzalez, Alet, Andersson, El-Kadi, Masters, Ewalds, Stott, Mohamed, Battaglia, Lam, and Willson]{Price2025}
Ilan Price, Alvaro Sanchez-Gonzalez, Ferran Alet, Tom~R. Andersson, Andrew El-Kadi, Dominic Masters, Timo Ewalds, Jacklynn Stott, Shakir Mohamed, Peter Battaglia, Remi Lam, and Matthew Willson.
\newblock {Probabilistic weather forecasting with machine learning}.
\newblock \emph{Nature}, 637:\penalty0 84--90, January 2025.
\newblock ISSN 1476-4687.
\newblock \doi{10.1038/s41586-024-08252-9}.

\bibitem[Yang et~al.(2025)Yang, Nai, Liu, Li, Chao, Wang, Wang, Li, Chen, Lu, Xiao, Boers, Yuan, and Pan]{Yang2025}
Shangshang Yang, Congyi Nai, Xinyan Liu, Weidong Li, Jie Chao, Jingnan Wang, Leyi Wang, Xichen Li, Xi~Chen, Bo~Lu, Ziniu Xiao, Niklas Boers, Huiling Yuan, and Baoxiang Pan.
\newblock {Generative assimilation and prediction for weather and climate}.
\newblock \emph{arXiv}, March 2025.
\newblock \doi{10.48550/arXiv.2503.03038}.

\bibitem[Maraun et~al.(2015)Maraun, Widmann, Gutiérrez, Kotlarski, Chandler, Hertig, Wibig, Huth, and Wilcke]{maraun2015VALUE}
Douglas Maraun, Martin Widmann, José~M. Gutiérrez, Sven Kotlarski, Richard~E. Chandler, Elke Hertig, Joanna Wibig, Radan Huth, and Renate~A.I. Wilcke.
\newblock Value: A framework to validate downscaling approaches for climate change studies.
\newblock \emph{Earth's Future}, 3\penalty0 (1):\penalty0 1--14, 2015.
\newblock \doi{https://doi.org/10.1002/2014EF000259}.
\newblock URL \url{https://agupubs.onlinelibrary.wiley.com/doi/abs/10.1002/2014EF000259}.

\bibitem[Dawid(1984)]{dawid1984pit}
A.~P. Dawid.
\newblock Present position and potential developments: Some personal views: Statistical theory: The prequential approach.
\newblock \emph{Journal of the Royal Statistical Society. Series A (General)}, 147\penalty0 (2):\penalty0 278--292, 1984.
\newblock ISSN 00359238, 23972327.
\newblock URL \url{http://www.jstor.org/stable/2981683}.

\bibitem[Volosciuk et~al.(2017)Volosciuk, Maraun, Vrac, and Widmann]{Volosciuk2017bcds}
C.~Volosciuk, D.~Maraun, M.~Vrac, and M.~Widmann.
\newblock A combined statistical bias correction and stochastic downscaling method for precipitation.
\newblock \emph{Hydrology and Earth System Sciences}, 21\penalty0 (3):\penalty0 1693--1719, 2017.
\newblock \doi{10.5194/hess-21-1693-2017}.
\newblock URL \url{https://hess.copernicus.org/articles/21/1693/2017/}.

\bibitem[Maraun(2013)]{maraun2013bias}
Douglas Maraun.
\newblock Bias correction, quantile mapping, and downscaling: Revisiting the inflation issue.
\newblock \emph{Journal of Climate}, 26\penalty0 (6):\penalty0 2137--2143, 2013.

\bibitem[Adewoyin et~al.(2021)Adewoyin, Dueben, Watson, He, and Dutta]{Adewoyin2021}
Rilwan~A. Adewoyin, Peter Dueben, Peter Watson, Yulan He, and Ritabrata Dutta.
\newblock {TRU-NET: a deep learning approach to high resolution prediction of rainfall}.
\newblock \emph{Machine Learning}, 110\penalty0 (8):\penalty0 2035--2062, 2021.
\newblock ISSN 1573-0565.

\bibitem[Harder et~al.(2023)Harder, Hernandez-Garcia, Ramesh, Yang, Sattegeri, Szwarcman, Watson, and Rolnick]{Harder2023}
Paula Harder, Alex Hernandez-Garcia, Venkatesh Ramesh, Qidong Yang, Prasanna Sattegeri, Daniela Szwarcman, Campbell Watson, and David Rolnick.
\newblock {Hard-Constrained Deep Learning for Climate Downscaling}.
\newblock \emph{Journal of Machine Learning Research}, 24\penalty0 (365):\penalty0 1--40, 2023.
\newblock ISSN 1533-7928.

\bibitem[Hess et~al.(2025)Hess, Aich, Pan, and Boers]{Hess2025}
Philipp Hess, Michael Aich, Baoxiang Pan, and Niklas Boers.
\newblock {Fast, scale-adaptive and uncertainty-aware downscaling of Earth system model fields with generative machine learning}.
\newblock \emph{Nat. Mach. Intell.}, 7:\penalty0 363--373, March 2025.
\newblock ISSN 2522-5839.
\newblock \doi{10.1038/s42256-025-00980-5}.

\bibitem[Bischoff and Deck(2024)]{Bischoff2024unpaireddownscaling}
Tobias Bischoff and Katherine Deck.
\newblock Unpaired downscaling of fluid flows with diffusion bridges.
\newblock \emph{Artificial Intelligence for the Earth Systems}, 3\penalty0 (2):\penalty0 e230039, 2024.

\bibitem[Schmidt and Ludwig(2024)]{schmidt2024wind}
Luca Schmidt and Nicole Ludwig.
\newblock Wind power assessment based on super-resolution and downscaling--a comparison of deep learning methods.
\newblock \emph{arXiv preprint arXiv:2407.08259}, 2024.

\bibitem[Rampal et~al.(2024)Rampal, Hobeichi, Gibson, Baño-Medina, Abramowitz, Beucler, González-Abad, Chapman, Harder, and Gutiérrez]{Rampal2024enhancing}
Neelesh Rampal, Sanaa Hobeichi, Peter~B. Gibson, Jorge Baño-Medina, Gab Abramowitz, Tom Beucler, Jose González-Abad, William Chapman, Paula Harder, and José~Manuel Gutiérrez.
\newblock Enhancing regional climate downscaling through advances in machine learning.
\newblock \emph{Artificial Intelligence for the Earth Systems}, 3\penalty0 (2):\penalty0 230066, 2024.
\newblock \doi{10.1175/AIES-D-23-0066.1}.

\bibitem[Groenke et~al.(2020)Groenke, Madaus, and Monteleoni]{groenke2020climalign}
Brian Groenke, Luke Madaus, and Claire Monteleoni.
\newblock Climalign: Unsupervised statistical downscaling of climate variables via normalizing flows.
\newblock In \emph{Proceedings of the 10th International Conference on Climate Informatics}, pages 60--66, 2020.

\bibitem[Harris et~al.(2022)Harris, McRae, Chantry, Dueben, and Palmer]{harris2022downscaling}
Lucy Harris, Andrew T.~T. McRae, Matthew Chantry, Peter~D. Dueben, and Tim~N. Palmer.
\newblock A generative deep learning approach to stochastic downscaling of precipitation forecasts.
\newblock \emph{Journal of Advances in Modeling Earth Systems}, 14\penalty0 (10), 2022.

\bibitem[Tomasi et~al.(2025)Tomasi, Franch, and Cristoforetti]{Tomasi2025}
Elena Tomasi, Gabriele Franch, and Marco Cristoforetti.
\newblock {Can AI be enabled to perform dynamical downscaling? A latent diffusion model to mimic kilometer-scale COSMO5.0{$\_$}CLM9 simulations}.
\newblock \emph{Geosci. Model Dev.}, 18\penalty0 (6):\penalty0 2051--2078, April 2025.
\newblock ISSN 1991-959X.
\newblock \doi{10.5194/gmd-18-2051-2025}.

\bibitem[Langguth et~al.(2024)Langguth, Patnala, Lehner, Dabernig, Mayer, Schicker, Austria, and Harder]{langguth2024}
Michael Langguth, Ankit Patnala, Sebastian Lehner, Markus Dabernig, Konrad Mayer, Irene Schicker, GeoSphere Austria, and Paula Harder.
\newblock A benchmark dataset for meteorological downscaling.
\newblock In \emph{International Conference on Learning Representations}, 2024.

\bibitem[Addison et~al.(2024)Addison, Kendon, Ravuri, Aitchison, and Watson]{addison2024machine}
Henry Addison, Elizabeth Kendon, Suman Ravuri, Laurence Aitchison, and Peter~AG Watson.
\newblock Machine learning emulation of precipitation from km-scale regional climate simulations using a diffusion model.
\newblock \emph{arXiv preprint arXiv:2407.14158}, 2024.

\bibitem[Zscheischler et~al.(2018)Zscheischler, Westra, van~den Hurk, Seneviratne, Ward, Pitman, AghaKouchak, Bresch, Leonard, Wahl, and Zhang]{Zscheischler2018}
Jakob Zscheischler, Seth Westra, Bart J. J.~M. van~den Hurk, Sonia~I. Seneviratne, Philip~J. Ward, Andy Pitman, Amir AghaKouchak, David~N. Bresch, Michael Leonard, Thomas Wahl, and Xuebin Zhang.
\newblock Future climate risk from compound events.
\newblock \emph{Nature Climate Change}, 8\penalty0 (6):\penalty0 469--477, 2018.
\newblock ISSN 1758-6798.
\newblock \doi{10.1038/s41558-018-0156-3}.
\newblock URL \url{https://doi.org/10.1038/s41558-018-0156-3}.

\bibitem[Srivastava et~al.(2024)Srivastava, Yang, Kerrigan, Dresdner, McGibbon, Bretherton, and Mandt]{srivastava2024precipitation}
Prakhar Srivastava, Ruihan Yang, Gavin Kerrigan, Gideon Dresdner, Jeremy~J McGibbon, Christopher~S. Bretherton, and Stephan Mandt.
\newblock Precipitation downscaling with spatiotemporal video diffusion.
\newblock In \emph{The Thirty-eighth Annual Conference on Neural Information Processing Systems}, 2024.
\newblock URL \url{https://openreview.net/forum?id=hhnkH8ex5d}.

\bibitem[Winkler et~al.(2024)Winkler, Harder, and Rolnick]{Winkler2024}
Christina Winkler, Paula Harder, and David Rolnick.
\newblock {Climate Variable Downscaling with Conditional Normalizing Flows}.
\newblock \emph{arXiv}, May 2024.

\bibitem[Bollmeyer et~al.(2015)Bollmeyer, Keller, Ohlwein, Wahl, Crewell, Friederichs, Hense, Keune, Kneifel, Pscheidt, Redl, and Steinke]{cosmo2015bollmeyer}
C.~Bollmeyer, J.~D. Keller, C.~Ohlwein, S.~Wahl, S.~Crewell, P.~Friederichs, A.~Hense, J.~Keune, S.~Kneifel, I.~Pscheidt, S.~Redl, and S.~Steinke.
\newblock Towards a high-resolution regional reanalysis for the european cordex domain.
\newblock \emph{Quarterly Journal of the Royal Meteorological Society}, 141\penalty0 (686):\penalty0 1--15, 2015.

\bibitem[M{\"u}ller et~al.(2018)M{\"u}ller, Jungclaus, Mauritsen, Baehr, Bittner, Budich, Bunzel, Esch, Ghosh, Haak, et~al.]{muller2018higher}
Wolfgang~A M{\"u}ller, Johann~H Jungclaus, Thorsten Mauritsen, Johanna Baehr, Matthias Bittner, R~Budich, Felix Bunzel, Monika Esch, Rohit Ghosh, Helmut Haak, et~al.
\newblock A higher-resolution version of the max planck institute earth system model (mpi-esm1. 2-hr).
\newblock \emph{Journal of Advances in Modeling Earth Systems}, 10\penalty0 (7):\penalty0 1383--1413, 2018.

\bibitem[Andrews et~al.(2020)Andrews, Ridley, Wood, Andrews, Blockley, Booth, Burke, Dittus, Florek, Gray, et~al.]{andrews2020historical}
Martin~B Andrews, Jeff~K Ridley, Richard~A Wood, Timothy Andrews, Edward~W Blockley, Ben Booth, Eleanor Burke, Andrea~J Dittus, Piotr Florek, Lesley~J Gray, et~al.
\newblock Historical simulations with hadgem3-gc3. 1 for cmip6.
\newblock \emph{Journal of Advances in Modeling Earth Systems}, 12\penalty0 (6):\penalty0 e2019MS001995, 2020.

\bibitem[Schwertfeger(2024)]{pythoncmethods}
Benjamin~T. Schwertfeger.
\newblock btschwertfeger/python-cmethods: v2.3.0, June 2024.
\newblock URL \url{https://doi.org/10.5281/zenodo.12168002}.

\bibitem[Gelman et~al.(2013)Gelman, Carlin, Stern, Dunson, Vehtari, and Rubin]{gelman2013bayesian}
Andrew Gelman, John~B. Carlin, Hal~S. Stern, David~B. Dunson, Aki Vehtari, and Donald~B. Rubin.
\newblock \emph{Bayesian Data Analysis}.
\newblock Chapman and Hall/CRC, 3 edition, 2013.
\newblock ISBN 978-1-4398-4095-5.

\bibitem[Bishop(2006)]{bishop2006pattern}
Christopher~M. Bishop.
\newblock \emph{Pattern Recognition and Machine Learning}.
\newblock Information Science and Statistics. Springer, 1 edition, 2006.
\newblock ISBN 978-0-387-31073-2.
\newblock Springer-Verlag New York; eBook packages: Computer Science (R0).

\bibitem[Sohl-Dickstein et~al.(2015)Sohl-Dickstein, Weiss, Maheswaranathan, and Ganguli]{dm2015sohldickstein}
Jascha Sohl-Dickstein, Eric Weiss, Niru Maheswaranathan, and Surya Ganguli.
\newblock Deep unsupervised learning using nonequilibrium thermodynamics.
\newblock In Francis Bach and David Blei, editors, \emph{Proceedings of the 32nd International Conference on Machine Learning}, volume~37 of \emph{Proceedings of Machine Learning Research}, pages 2256--2265, Lille, France, 07--09 Jul 2015. PMLR.

\bibitem[Kingma et~al.(2021)Kingma, Salimans, Poole, and Ho]{variationaldms2021kingma}
Diederik Kingma, Tim Salimans, Ben Poole, and Jonathan Ho.
\newblock Variational diffusion models.
\newblock In M.~Ranzato, A.~Beygelzimer, Y.~Dauphin, P.S. Liang, and J.~Wortman Vaughan, editors, \emph{Advances in Neural Information Processing Systems}, volume~34, pages 21696--21707. Curran Associates, Inc., 2021.

\bibitem[Karras et~al.(2022)Karras, Aittala, Aila, and Laine]{edm2022karras}
Tero Karras, Miika Aittala, Timo Aila, and Samuli Laine.
\newblock Elucidating the design space of diffusion-based generative models.
\newblock In S.~Koyejo, S.~Mohamed, A.~Agarwal, D.~Belgrave, K.~Cho, and A.~Oh, editors, \emph{Advances in Neural Information Processing Systems}, volume~35, pages 26565--26577. Curran Associates, Inc., 2022.

\bibitem[S\"arkk\"a and Solin(2019)]{sdebook2019sarkka}
S.~S\"arkk\"a and A.~Solin.
\newblock \emph{Applied Stochastic Differential Equations}.
\newblock Cambridge University Press, 2019.

\bibitem[Anderson(1982)]{reversediff1982anderson}
Brian~D.O. Anderson.
\newblock Reverse-time diffusion equation models.
\newblock \emph{Stochastic Processes and their Applications}, 12\penalty0 (3):\penalty0 313--326, 1982.
\newblock ISSN 0304-4149.

\bibitem[Zhang and Chen(2023)]{zhang2023fast}
Qinsheng Zhang and Yongxin Chen.
\newblock Fast sampling of diffusion models with exponential integrator.
\newblock In \emph{The Eleventh International Conference on Learning Representations}, 2023.

\bibitem[Song et~al.(2023)Song, Dhariwal, Chen, and Sutskever]{consistency2023song}
Yang Song, Prafulla Dhariwal, Mark Chen, and Ilya Sutskever.
\newblock Consistency models.
\newblock \emph{arXiv preprint arXiv:2303.01469}, 2023.

\bibitem[Zheng et~al.(2023)Zheng, Nie, Vahdat, Azizzadenesheli, and Anandkumar]{pmlr-v202-zheng23d}
Hongkai Zheng, Weili Nie, Arash Vahdat, Kamyar Azizzadenesheli, and Anima Anandkumar.
\newblock Fast sampling of diffusion models via operator learning.
\newblock In Andreas Krause, Emma Brunskill, Kyunghyun Cho, Barbara Engelhardt, Sivan Sabato, and Jonathan Scarlett, editors, \emph{Proceedings of the 40th International Conference on Machine Learning}, volume 202 of \emph{Proceedings of Machine Learning Research}, pages 42390--42402. PMLR, 23--29 Jul 2023.

\bibitem[Vincent(2011)]{denoisingsm2011vincent}
Pascal Vincent.
\newblock A connection between score matching and denoising autoencoders.
\newblock \emph{Neural Computation}, 23\penalty0 (7):\penalty0 1661--1674, 2011.

\bibitem[Ronneberger et~al.(2015)Ronneberger, P.Fischer, and Brox]{unet2015ronneberger}
O.~Ronneberger, P.Fischer, and T.~Brox.
\newblock U-net: Convolutional networks for biomedical image segmentation.
\newblock In \emph{Medical Image Computing and Computer-Assisted Intervention (MICCAI)}, volume 9351 of \emph{LNCS}, pages 234--241. Springer, 2015.

\bibitem[Chung et~al.(2023)Chung, Kim, Mccann, Klasky, and Ye]{dps2023chung}
Hyungjin Chung, Jeongsol Kim, Michael~Thompson Mccann, Marc~Louis Klasky, and Jong~Chul Ye.
\newblock Diffusion posterior sampling for general noisy inverse problems.
\newblock In \emph{The Eleventh International Conference on Learning Representations}, 2023.

\bibitem[{Met Office}(2010 - 2015)]{Cartopy}
{Met Office}.
\newblock \emph{Cartopy: a cartographic python library with a Matplotlib interface}.
\newblock Exeter, Devon, 2010 - 2015.
\newblock URL \url{http://scitools.org.uk/cartopy}.

\bibitem[Paszke et~al.(2019)Paszke, Gross, Massa, Lerer, Bradbury, Chanan, Killeen, Lin, Gimelshein, Antiga, Desmaison, Köpf, Yang, DeVito, Raison, Tejani, Chilamkurthy, Steiner, Fang, Bai, and Chintala]{pytorch}
Adam Paszke, Sam Gross, Francisco Massa, Adam Lerer, James Bradbury, Gregory Chanan, Trevor Killeen, Zeming Lin, Natalia Gimelshein, Luca Antiga, Alban Desmaison, Andreas Köpf, Edward Yang, Zach DeVito, Martin Raison, Alykhan Tejani, Sasank Chilamkurthy, Benoit Steiner, Lu~Fang, Junjie Bai, and Soumith Chintala.
\newblock Pytorch: An imperative style, high-performance deep learning library, 2019.
\newblock URL \url{https://arxiv.org/abs/1912.01703}.

\bibitem[Ruzanski and Chandrasekar(2011)]{ruzanski2011scale}
Evan Ruzanski and V~Chandrasekar.
\newblock Scale filtering for improved nowcasting performance in a high-resolution x-band radar network.
\newblock \emph{IEEE transactions on geoscience and remote sensing}, 49\penalty0 (6):\penalty0 2296--2307, 2011.

\bibitem[Hess et~al.(2023)Hess, Lange, Schötz, and Boers]{hess2023bc}
Philipp Hess, Stefan Lange, Christof Schötz, and Niklas Boers.
\newblock Deep learning for bias-correcting cmip6-class earth system models.
\newblock \emph{Earth's Future}, 11\penalty0 (10):\penalty0 e2023EF004002, 2023.

\bibitem[Pulkkinen et~al.(2019)Pulkkinen, Nerini, P\'erez~Hortal, Velasco-Forero, Seed, Germann, and Foresti]{pulkkinen2019pysteps}
S.~Pulkkinen, D.~Nerini, A.~A. P\'erez~Hortal, C.~Velasco-Forero, A.~Seed, U.~Germann, and L.~Foresti.
\newblock Pysteps: an open-source python library for probabilistic precipitation nowcasting (v1.0).
\newblock \emph{Geoscientific Model Development}, 12\penalty0 (10):\penalty0 4185--4219, 2019.

\bibitem[Haas et~al.(2024)Haas, Krien, Schachler, Bot, Zeli, Maurer, Shivam, Witte, Rasti, Seth, and Bosch]{haas2024windpowerlib}
Sabine Haas, Uwe Krien, Birgit Schachler, Stickler Bot, Velibor Zeli, Florian Maurer, Kumar Shivam, Francesco Witte, Sasan~Jacob Rasti, Seth, and Stephen Bosch.
\newblock wind-python/windpowerlib: Update release, February 2024.
\newblock URL \url{https://doi.org/10.5281/zenodo.10685057}.

\bibitem[Carrillo et~al.(2013)Carrillo, {Obando Montaño}, Cidrás, and Díaz-Dorado]{carrillo2013windpower}
C.~Carrillo, A.F. {Obando Montaño}, J.~Cidrás, and E.~Díaz-Dorado.
\newblock Review of power curve modelling for wind turbines.
\newblock \emph{Renewable and Sustainable Energy Reviews}, 21:\penalty0 572--581, 2013.
\newblock ISSN 1364-0321.

\bibitem[Vandal et~al.(2019)Vandal, Kodra, and Ganguly]{vandal2019intercomparison}
Thomas Vandal, Evan Kodra, and Auroop~R Ganguly.
\newblock Intercomparison of machine learning methods for statistical downscaling: the case of daily and extreme precipitation.
\newblock \emph{Theoretical and Applied Climatology}, 137:\penalty0 557--570, 2019.

\bibitem[Bischoff et~al.(2024)Bischoff, Darcher, Deistler, Gao, Gerken, Gloeckler, Haxel, Kapoor, Lappalainen, Macke, et~al.]{bischoff2024practical}
Sebastian Bischoff, Alana Darcher, Michael Deistler, Richard Gao, Franziska Gerken, Manuel Gloeckler, Lisa Haxel, Jaivardhan Kapoor, Janne~K Lappalainen, Jakob~H Macke, et~al.
\newblock A practical guide to sample-based statistical distances for evaluating generative models in science.
\newblock \emph{Transactions on Machine Learning Research}, 2024.

\bibitem[Wang et~al.(2003)Wang, Simoncelli, and Bovik]{wang2003ssim}
Z.~Wang, E.P. Simoncelli, and A.C. Bovik.
\newblock Multiscale structural similarity for image quality assessment.
\newblock In \emph{The Thrity-Seventh Asilomar Conference on Signals, Systems \& Computers, 2003}, volume~2, pages 1398--1402 Vol.2, 2003.

\bibitem[Feller(1949)]{feller1949theory}
W~Feller.
\newblock On the theory of stochastic processes, with particular reference to applications.
\newblock In \emph{First Berkeley Symposium on Mathematical Statistics and Probability}, pages 403--432, 1949.

\end{thebibliography}


\begin{thebibliography}{18}
\providecommand{\natexlab}[1]{#1}
\providecommand{\url}[1]{\texttt{#1}}
\expandafter\ifx\csname urlstyle\endcsname\relax
  \providecommand{\doi}[1]{doi: #1}\else
  \providecommand{\doi}{doi: \begingroup \urlstyle{rm}\Url}\fi

\bibitem[Ruzanski and Chandrasekar(2011)]{ruzanski2011scale}
Evan Ruzanski and V~Chandrasekar.
\newblock Scale filtering for improved nowcasting performance in a high-resolution x-band radar network.
\newblock \emph{IEEE transactions on geoscience and remote sensing}, 49\penalty0 (6):\penalty0 2296--2307, 2011.

\bibitem[Harris et~al.(2022)Harris, McRae, Chantry, Dueben, and Palmer]{harris2022downscaling}
Lucy Harris, Andrew T.~T. McRae, Matthew Chantry, Peter~D. Dueben, and Tim~N. Palmer.
\newblock A generative deep learning approach to stochastic downscaling of precipitation forecasts.
\newblock \emph{Journal of Advances in Modeling Earth Systems}, 14\penalty0 (10), 2022.

\bibitem[Hess et~al.(2023)Hess, Lange, Schötz, and Boers]{hess2023bc}
Philipp Hess, Stefan Lange, Christof Schötz, and Niklas Boers.
\newblock Deep learning for bias-correcting cmip6-class earth system models.
\newblock \emph{Earth's Future}, 11\penalty0 (10):\penalty0 e2023EF004002, 2023.

\bibitem[Pulkkinen et~al.(2019)Pulkkinen, Nerini, P\'erez~Hortal, Velasco-Forero, Seed, Germann, and Foresti]{pulkkinen2019pysteps}
S.~Pulkkinen, D.~Nerini, A.~A. P\'erez~Hortal, C.~Velasco-Forero, A.~Seed, U.~Germann, and L.~Foresti.
\newblock Pysteps: an open-source python library for probabilistic precipitation nowcasting (v1.0).
\newblock \emph{Geoscientific Model Development}, 12\penalty0 (10):\penalty0 4185--4219, 2019.

\bibitem[Haas et~al.(2024)Haas, Krien, Schachler, Bot, Zeli, Maurer, Shivam, Witte, Rasti, Seth, and Bosch]{haas2024windpowerlib}
Sabine Haas, Uwe Krien, Birgit Schachler, Stickler Bot, Velibor Zeli, Florian Maurer, Kumar Shivam, Francesco Witte, Sasan~Jacob Rasti, Seth, and Stephen Bosch.
\newblock wind-python/windpowerlib: Update release, February 2024.
\newblock URL \url{https://doi.org/10.5281/zenodo.10685057}.

\bibitem[Carrillo et~al.(2013)Carrillo, {Obando Montaño}, Cidrás, and Díaz-Dorado]{carrillo2013windpower}
C.~Carrillo, A.F. {Obando Montaño}, J.~Cidrás, and E.~Díaz-Dorado.
\newblock Review of power curve modelling for wind turbines.
\newblock \emph{Renewable and Sustainable Energy Reviews}, 21:\penalty0 572--581, 2013.
\newblock ISSN 1364-0321.

\bibitem[Vandal et~al.(2019)Vandal, Kodra, and Ganguly]{vandal2019intercomparison}
Thomas Vandal, Evan Kodra, and Auroop~R Ganguly.
\newblock Intercomparison of machine learning methods for statistical downscaling: the case of daily and extreme precipitation.
\newblock \emph{Theoretical and Applied Climatology}, 137:\penalty0 557--570, 2019.

\bibitem[Song et~al.(2021{\natexlab{a}})Song, Meng, and Ermon]{ddim2021song}
Jiaming Song, Chenlin Meng, and Stefano Ermon.
\newblock Denoising diffusion implicit models.
\newblock In \emph{International Conference on Learning Representations}, 2021{\natexlab{a}}.

\bibitem[Schmidt and Ludwig(2024)]{schmidt2024wind}
Luca Schmidt and Nicole Ludwig.
\newblock Wind power assessment based on super-resolution and downscaling--a comparison of deep learning methods.
\newblock \emph{arXiv preprint arXiv:2407.08259}, 2024.

\bibitem[Bischoff et~al.(2024)Bischoff, Darcher, Deistler, Gao, Gerken, Gloeckler, Haxel, Kapoor, Lappalainen, Macke, et~al.]{bischoff2024practical}
Sebastian Bischoff, Alana Darcher, Michael Deistler, Richard Gao, Franziska Gerken, Manuel Gloeckler, Lisa Haxel, Jaivardhan Kapoor, Janne~K Lappalainen, Jakob~H Macke, et~al.
\newblock A practical guide to sample-based statistical distances for evaluating generative models in science.
\newblock \emph{Transactions on Machine Learning Research}, 2024.

\bibitem[Wang et~al.(2003)Wang, Simoncelli, and Bovik]{wang2003ssim}
Z.~Wang, E.P. Simoncelli, and A.C. Bovik.
\newblock Multiscale structural similarity for image quality assessment.
\newblock In \emph{The Thrity-Seventh Asilomar Conference on Signals, Systems \& Computers, 2003}, volume~2, pages 1398--1402 Vol.2, 2003.

\bibitem[Chung et~al.(2023)Chung, Kim, Mccann, Klasky, and Ye]{dps2023chung}
Hyungjin Chung, Jeongsol Kim, Michael~Thompson Mccann, Marc~Louis Klasky, and Jong~Chul Ye.
\newblock Diffusion posterior sampling for general noisy inverse problems.
\newblock In \emph{The Eleventh International Conference on Learning Representations}, 2023.

\bibitem[S\"arkk\"a and Solin(2019)]{sdebook2019sarkka}
S.~S\"arkk\"a and A.~Solin.
\newblock \emph{Applied Stochastic Differential Equations}.
\newblock Cambridge University Press, 2019.

\bibitem[Ho et~al.(2020)Ho, Jain, and Abbeel]{ddpm2020ho}
Jonathan Ho, Ajay Jain, and Pieter Abbeel.
\newblock Denoising diffusion probabilistic models.
\newblock In H.~Larochelle, M.~Ranzato, R.~Hadsell, M.F. Balcan, and H.~Lin, editors, \emph{Advances in Neural Information Processing Systems}, volume~33, pages 6840--6851. Curran Associates, Inc., 2020.

\bibitem[Song et~al.(2021{\natexlab{b}})Song, Sohl-Dickstein, Kingma, Kumar, Ermon, and Poole]{sdedm2021song}
Yang Song, Jascha Sohl-Dickstein, Diederik~P Kingma, Abhishek Kumar, Stefano Ermon, and Ben Poole.
\newblock Score-based generative modeling through stochastic differential equations.
\newblock In \emph{International Conference on Learning Representations}, 2021{\natexlab{b}}.

\bibitem[Feller(1949)]{feller1949theory}
W~Feller.
\newblock On the theory of stochastic processes, with particular reference to applications.
\newblock In \emph{First Berkeley Symposium on Mathematical Statistics and Probability}, pages 403--432, 1949.

\bibitem[Sohl-Dickstein et~al.(2015)Sohl-Dickstein, Weiss, Maheswaranathan, and Ganguli]{dm2015sohldickstein}
Jascha Sohl-Dickstein, Eric Weiss, Niru Maheswaranathan, and Surya Ganguli.
\newblock Deep unsupervised learning using nonequilibrium thermodynamics.
\newblock In Francis Bach and David Blei, editors, \emph{Proceedings of the 32nd International Conference on Machine Learning}, volume~37 of \emph{Proceedings of Machine Learning Research}, pages 2256--2265, Lille, France, 07--09 Jul 2015. PMLR.

\bibitem[Kingma et~al.(2021)Kingma, Salimans, Poole, and Ho]{variationaldms2021kingma}
Diederik Kingma, Tim Salimans, Ben Poole, and Jonathan Ho.
\newblock Variational diffusion models.
\newblock In M.~Ranzato, A.~Beygelzimer, Y.~Dauphin, P.S. Liang, and J.~Wortman Vaughan, editors, \emph{Advances in Neural Information Processing Systems}, volume~34, pages 21696--21707. Curran Associates, Inc., 2021.

\end{thebibliography}

\cleardoublepage

\renewcommand{\figurename}{Supplementary Figure}
\renewcommand{\tablename}{Supplementary Table}
\setcounter{figure}{0}
\setcounter{table}{0}

\renewcommand{\thesection}{Supplementary Section \arabic{section}}

\crefname{figure}{Supplementary Figure}{Supplementary Figures}
\crefname{table}{Supplementary Table}{Supplementary Tables}

\section{Spatial region: coarse and fine grid}

\begin{figure}[h!]
    \centering
    \includegraphics[width=\linewidth]{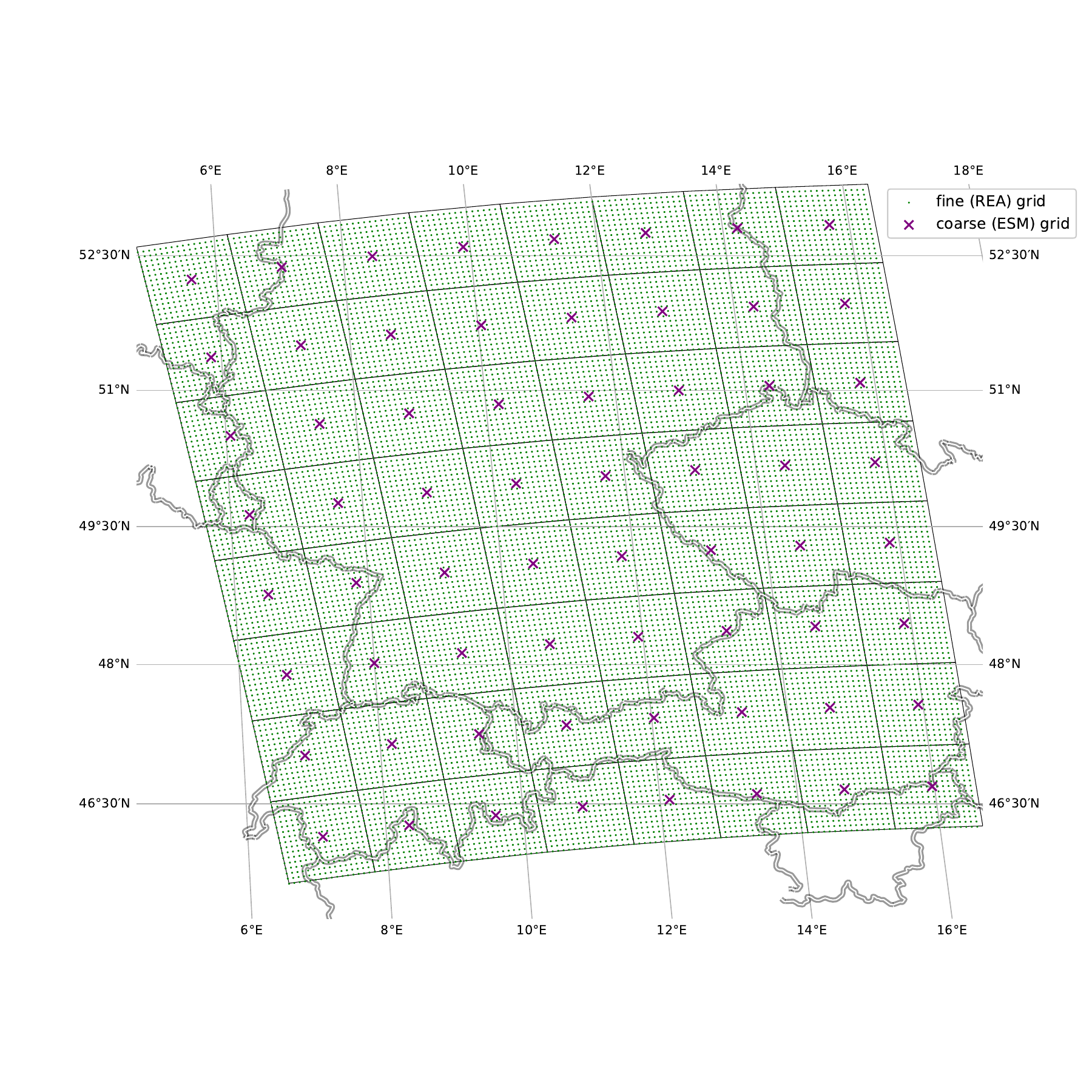}
    \caption{This plot shows the spatial region considered in this study.
    The coarse $8\times 8$-node grid is marked with purple crosses.
    The fine $128\times 128$-node grid is marked with green dots.
    Each coarse-grid node lies in the center of a corresponding $16\times 16$-patch of high-resolution grid nodes.
    Both grids span exactly the same area.
    }
    \label{fig:grid-points}
\end{figure}

\cleardoublepage

\section{Spatial patterns on multiple scales}\label{subsec:supp:rapsd}

\begin{figure}[h!]
     \centering
     \begin{subfigure}[b]{\linewidth}
         \centering
         \includegraphics[width=\linewidth]{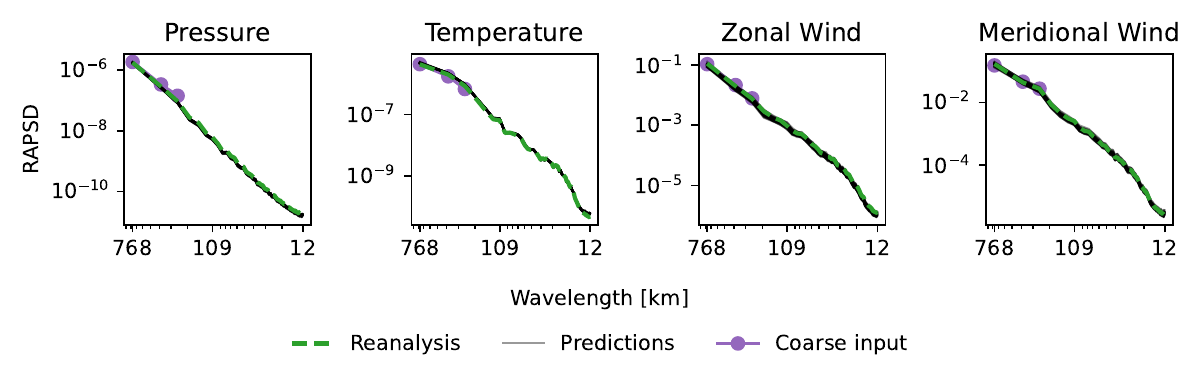}
         \caption{RAPSD for reanalysis data, coarse input, and predicted fine-grained reanalysis data.}
         \label{fig:supp:rapsd-storm}
     \end{subfigure}
     \begin{subfigure}[b]{\linewidth}
         \centering
         \includegraphics[width=\linewidth]{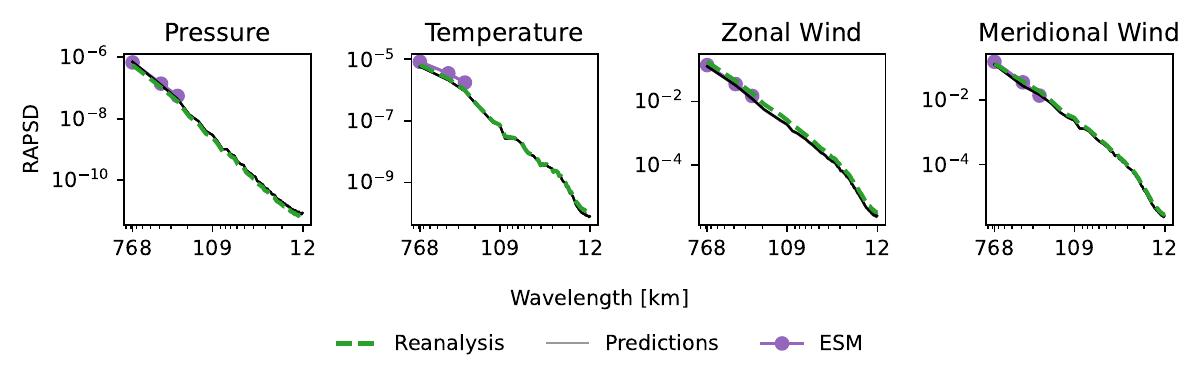}
         \caption{RAPSD for reanalysis data, CMIP6 simulations, and downscaled CMIP6 simulations.}
         \label{fig:supp:rapsd-clim}
     \end{subfigure}
     \caption{
     This plot shows the RAPSD for reanalysis data, coarse input, and the corresponding downscaled predictions.
     Subplot (a) mirrors the experimental setup of the on-model experiment that predicts high-resolution reanalysis data during the cyclone "Friederike" in January 2018. Subplot (b) covers the CMIP6 downscaling setup as described in the Methods section.
     }
     \label{fig:supp:rapsd}
\end{figure}

To assess the performance of the downscaling model across multiple spatial length scales, we show that the radially averaged power spectral densities (RAPSD) \citep{ruzanski2011scale} of the predictions align with the reanalysis data in \cref{fig:supp:rapsd}.
The RAPSD is computed by averaging the power spectrum over all directions of the same wavenumber in Fourier space.
The quantity is commonly used in the context of weather dynamics, especially when estimating precipitation \citep[for example]{harris2022downscaling,hess2023bc}.
We use the open-source \texttt{pysteps} package by \citet{pulkkinen2019pysteps} to compute the RAPSD.
The reported RAPSD values are averages over the considered time period.

\cleardoublepage

\section{Embedding the predicted region in an extended spatial context}

\begin{figure}[h!]
    \centering
    \includegraphics[width=\linewidth]{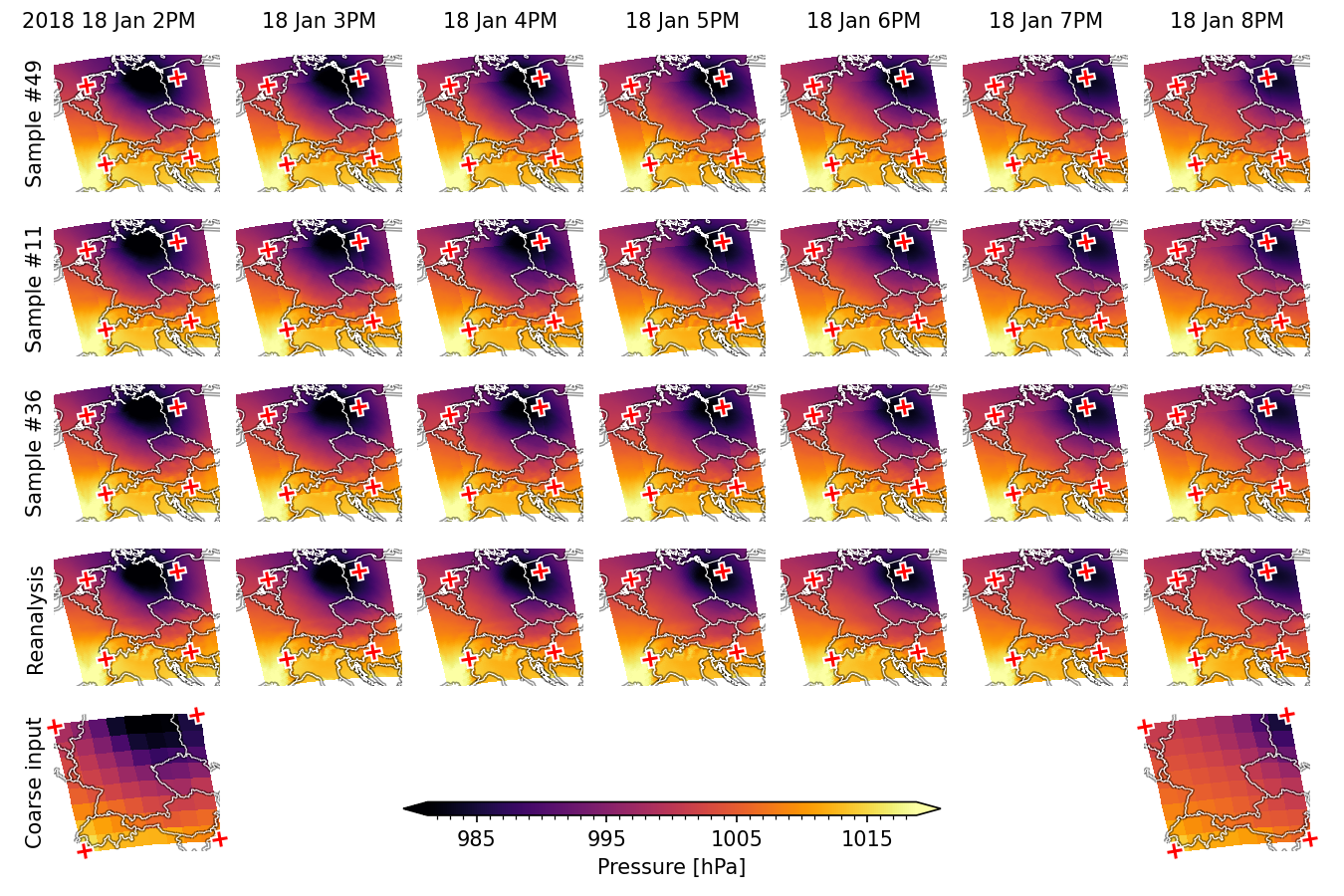}
    \caption{Predicting mean sea-level pressure during the cyclone "Friederike" (c.f. experiment above). Outside of the prediction range, which lies in the region indicated by red crosses, reanalysis data is filled in to give an impression as to how the predictions fit into more global dynamics.}
    \label{fig:supp:psl-big-grid}
\end{figure}

\begin{figure}[h!]
    \centering
    \includegraphics[width=\linewidth]{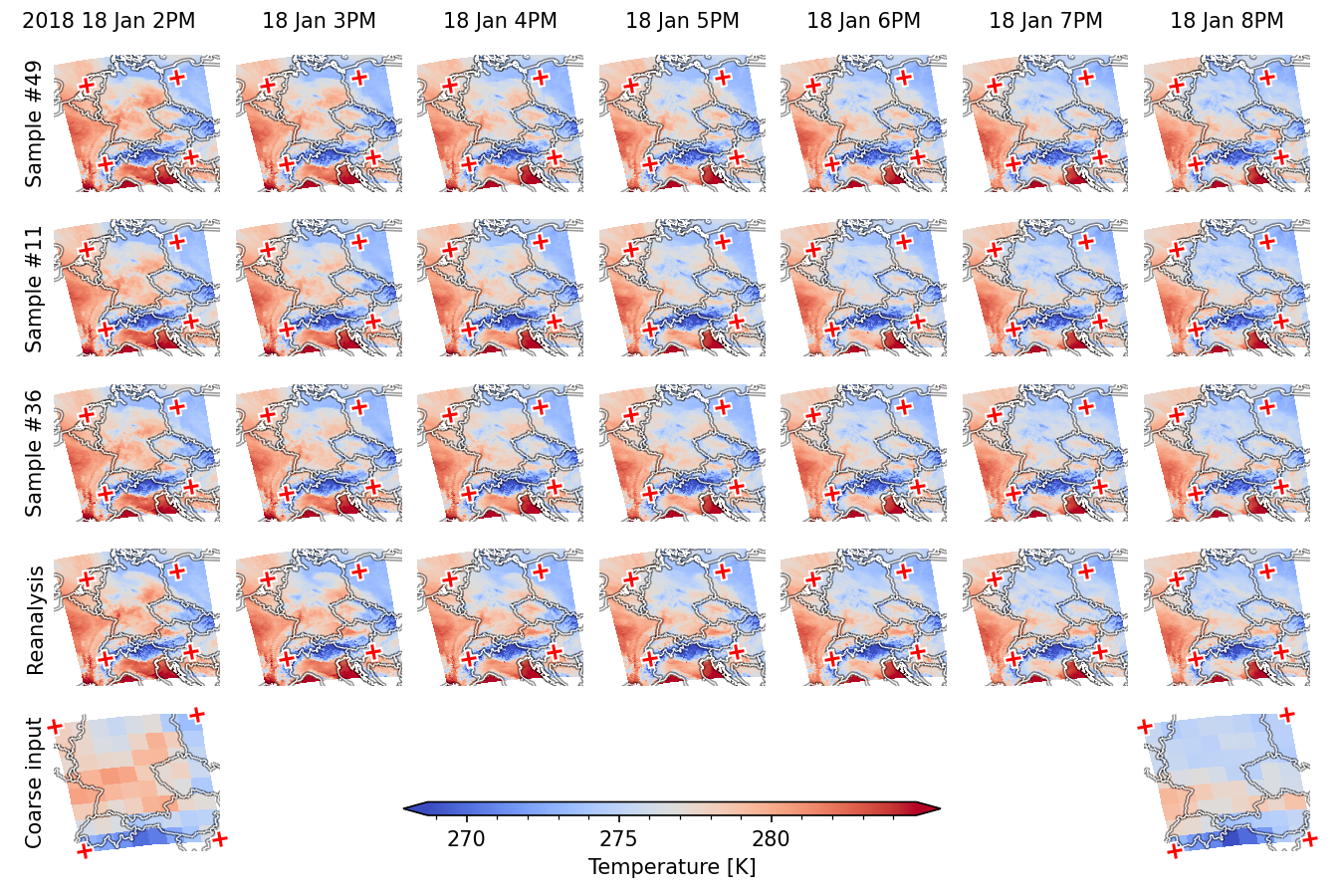}
    \caption{Predicting surface air temperature during the cyclone "Friederike" (c.f. experiment above). Outside of the prediction range, which lies in the region indicated by red crosses, reanalysis data is filled in to give an impression as to how the predictions fit into more global dynamics.}
    \label{fig:supp:tas-big-grid}
\end{figure}

\begin{figure}[h!]
    \centering
    \includegraphics[width=\linewidth]{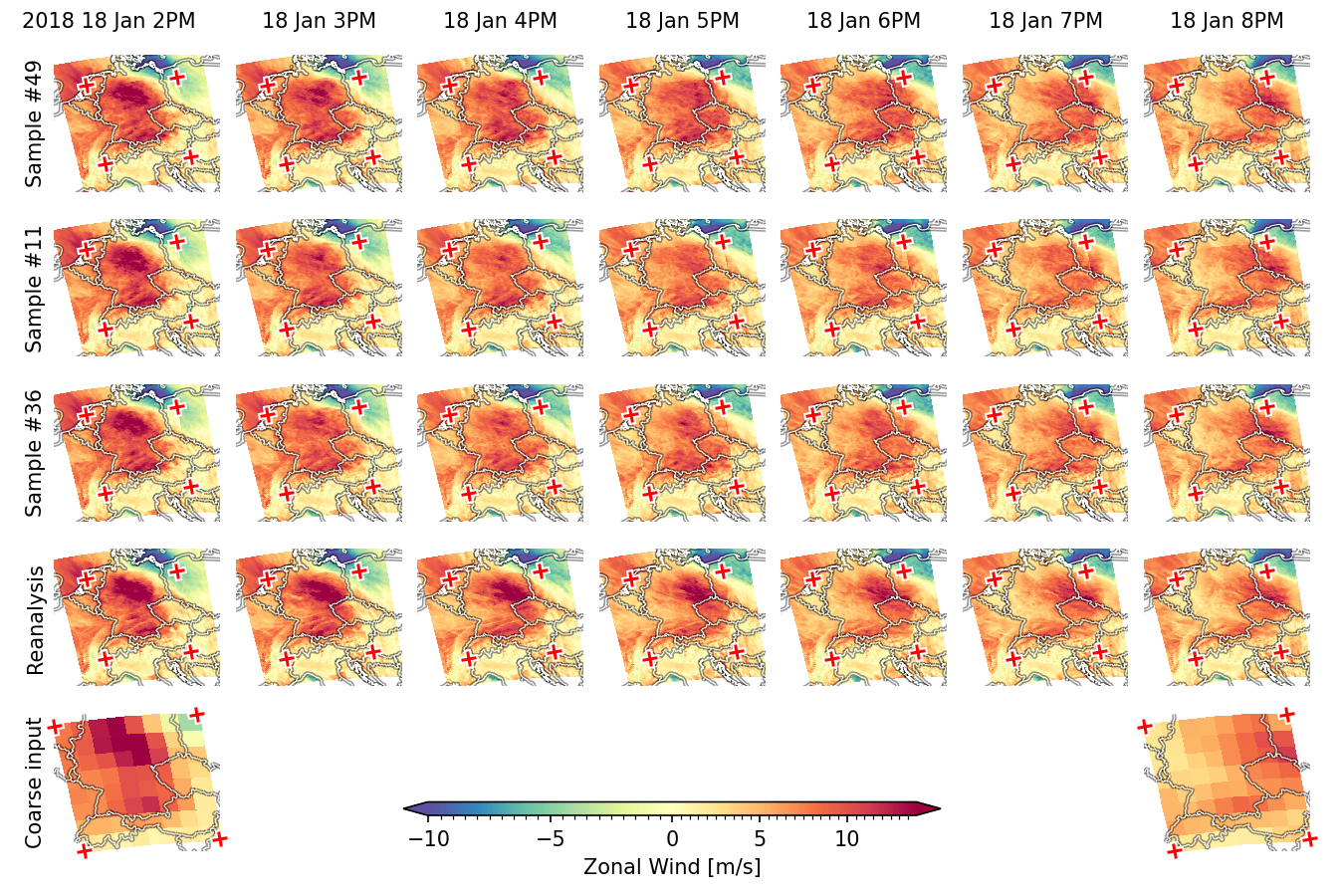}
    \caption{Predicting zonal wind speed during the cyclone "Friederike" (c.f. experiment above). Outside of the prediction range, which lies in the region indicated by red crosses, reanalysis data is filled in to give an impression as to how the predictions fit into more global dynamics.}
    \label{fig:supp:uas-big-grid}
\end{figure}

\begin{figure}[h!]
    \centering
    \includegraphics[width=\linewidth]{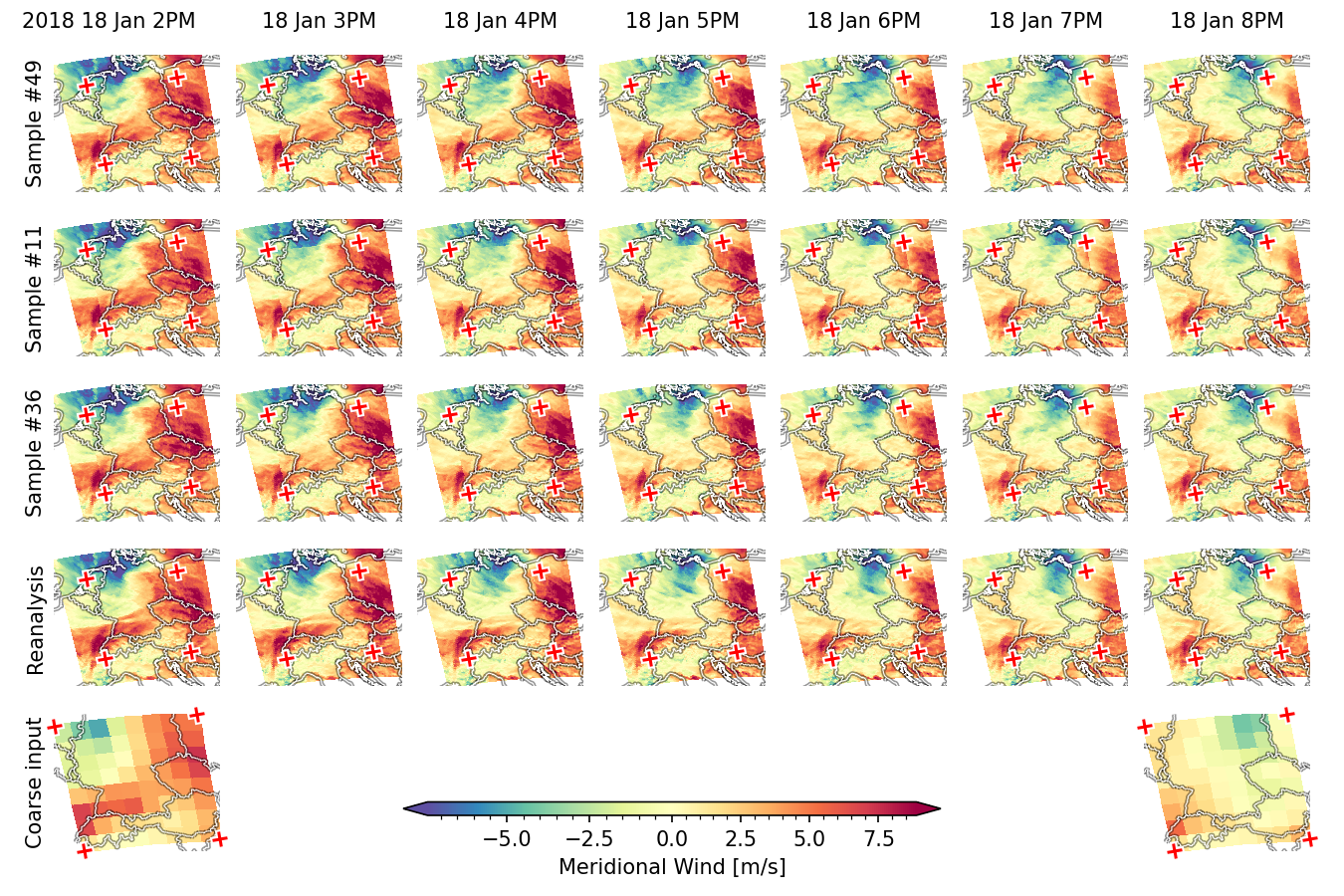}
    \caption{Predicting meridional wind speed during the cyclone "Friederike" (c.f. experiment above). Outside of the prediction range, which lies in the region indicated by red crosses, reanalysis data is filled in to give an impression as to how the predictions fit into more global dynamics.}
    \label{fig:supp:vas-big-grid}
\end{figure}

We embed the high-resolution predictions during the cyclone "Friederike" (c.f.~Figure 4) into a larger spatial context in order to learn about how long-distance interconnections between the studied spatial region and its surroundings are captured by the model.
We argue that it is likely that our statistical downscaling model is able to capture the global connectedness of weather dynamics that are contained in the reanalysis data it is trained on.
Concretely, \crefrange{fig:supp:psl-big-grid}{fig:supp:vas-big-grid} visualize the spatiotemporal dynamics of downscaled samples, reanalysis data, and conditioning input. Thereby, the spatial region is extended beyond the one considered in this study. Reanalysis data is used to fill in the regions outside of the predicted patch.
Clearly visible or implausible transitions at the edges of the patch would indicate that the model predictions do not align with the surrounding spatial context.
We find, however, that there is a smooth transition from outside the predicted area to its interior for the generated high-resolution predictions. The generated local dynamics seamlessly fit into the more surrounding context of reanalysis data.

\cleardoublepage

\section{Wind power prediction}\label{supp:subsec:windpower}

\begin{figure}[h!]
    \centering
    \includegraphics[width=\linewidth]{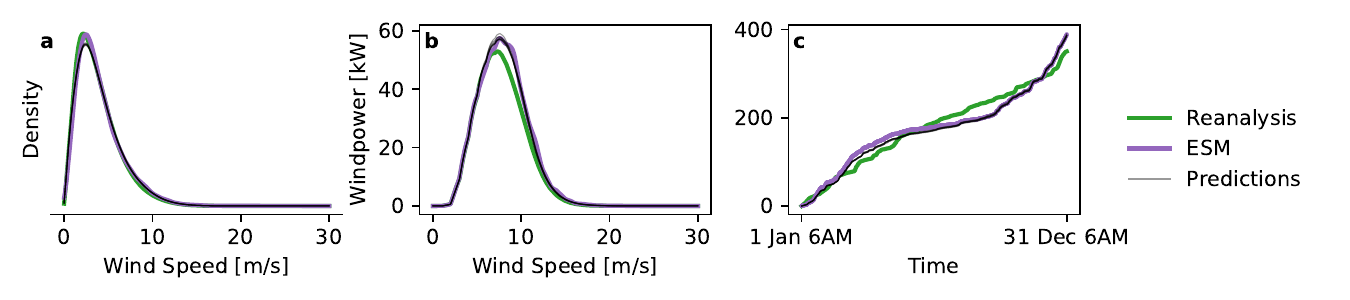}
    \caption{\textbf{Estimated generated wind power over time, averaged over the spatial domain.}
    This plot analyzes the local wind-speed predictions via the generated wind-power derived from the wind speeds.
    The full spatial region is considered over the year 2014.
    Subplot \labelize{a} visualizes a kernel-density estimate of the aggregated wind speeds for reanalysis data (green), ESM simulations (purple), and downscaled predictions (black).
    In \labelize{b}, we plot the generated wind power for the entire range of wind speeds from $0$m/s to $30$m/s, weighted by the density of the estimated wind speeds from \labelize{a}. We use a wind-power curve of the form \citet[Fig. 1]{carrillo2013windpower}.
    This estimates the wind power that is generated for the respective wind speeds and accounts for the frequency at which these wind speeds occur.
    Finally, subplot \labelize{c} shows the cumulative sum of the generated wind powers over time.
    All wind-power values are normalized by the number of time steps on the respective temporal grids to align the wind-power scales.
    From the black lines aligning closely with the purple line, we conclude that the downscaled predictions preserve the aggregated wind-power generation as simulated by the ESM.
    }
    \label{fig:supp:windpower-clim}
\end{figure}

\begin{figure}[h!]
     \centering
     \begin{subfigure}[b]{.9\linewidth}
         \centering
         \includegraphics[width=\linewidth]{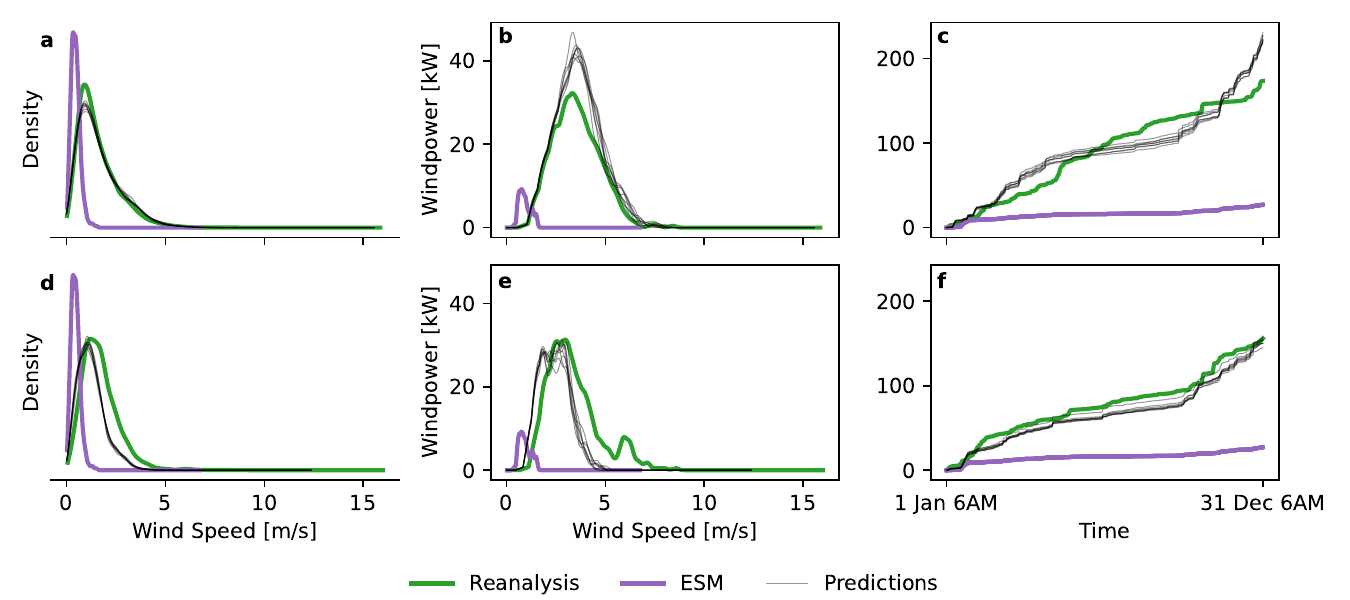}
         \caption{Compare generated windpower at two distinct locations that share a single node on the coarse ESM grid.
         In this instance, the ESM underpredicts the generated windpower.
         }
         \label{fig:supp:windpower-clim-loc1}
     \end{subfigure}
     \begin{subfigure}[b]{.9\linewidth}
         \centering
         \includegraphics[width=\linewidth]{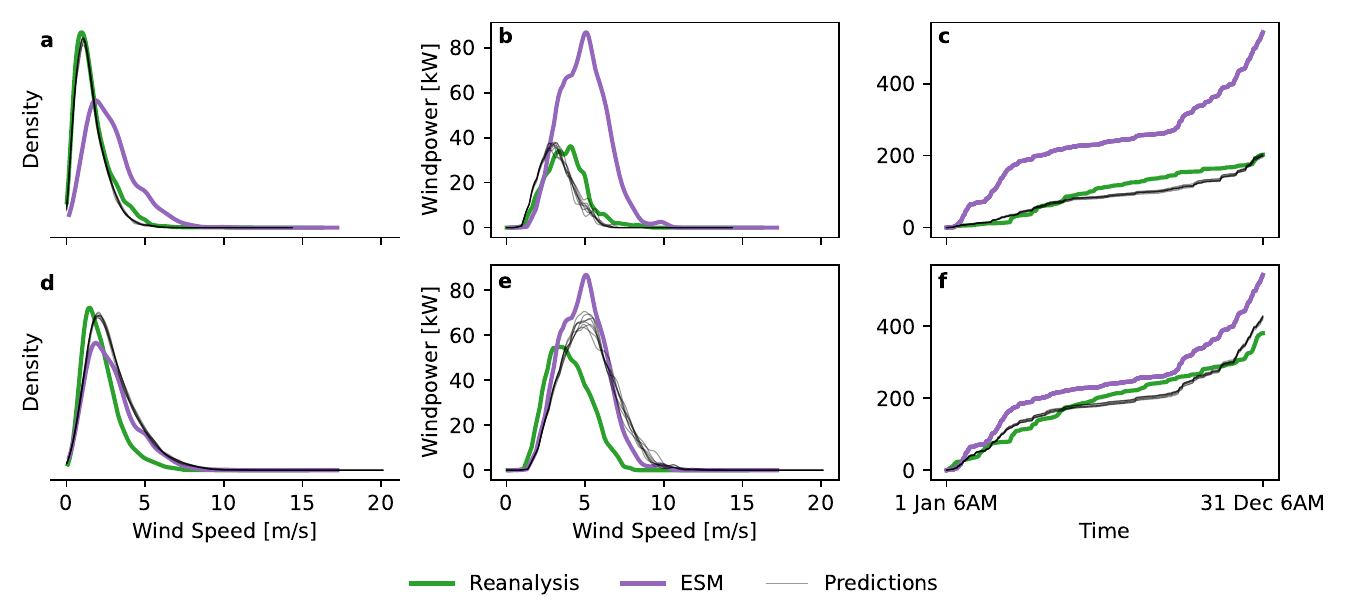}
         \caption{Compare generated windpower at two distinct locations that share a single node on the coarse ESM grid.
         In this instance, the ESM overpredicts the generated windpower.
         }
         \label{fig:supp:windpower-clim-loc2}
     \end{subfigure}
     \caption{\textbf{Local estimates of wind power over time.}
     This plot investigates two pairs of locations that are each constrained by a single spatial node on the coarse climate grid.
     The layout of each row of the subplots (a) and (b) mirrors \cref{fig:supp:windpower-clim}; each row corresponds to a single location on the fine-resolution grid, respectively.
     }
     \label{fig:supp:windpower-clim-local}
\end{figure}

A crucial motivation for spatiotemporal downscaling climate simulations to the weather scale is downstream tasks that require future local weather patterns.
We provide an exemplary evaluation of the estimated generated wind powers as computed from the CMIP6 simulation, the downscaled predictions, and the reanalysis data.
We find that, when computing the spatial average of generated wind powers, the wind-speed and wind-power predictions of the ESM are matched by the downscaled samples in distribution (\cref{fig:supp:windpower-clim}).
Comparing single locations reveals that the ESM locally sometimes over-predicts or under-predicts the wind power generated following the reanalysis data. For both cases, we consistently find that our model corrects the respective over- and under-estimation for multiple randomly selected locations.
We compare estimated densities of the wind-speed values (\cref{fig:supp:windpower-clim} \labelize{a}) and use this distribution to derive the amount of wind power generated from the respectively predicted wind speeds (\cref{fig:supp:windpower-clim} \labelize{b}).
For that, we first compute the wind-power curve for a range of wind-speed values ($0$m/s to $30$m/s), using the open-source package \texttt{windpowerlib} \citep{haas2024windpowerlib} and an arbitrary wind-turbine model (turbine type: "E-115/3000", hub-height: $100$m).
Then, we weigh this power curve with the density of predicted wind speeds to obtain the predicted wind power for each wind speed, taking into account how likely this wind speed value is to occur in the predictions.
Finally, we compare the accumulated generated wind power over time.
We show the comparison between two pairs of locations in \cref{fig:supp:windpower-clim-local}.


\cleardoublepage

\section{Quantitative evaluation}

We first apply quantile-mapping to bias-correct the climate patches. These frames are used as input for the different downscaling algorithms.
Our model is compared to the following two benchmark approaches.

\paragraph{Benchmark 1: Interpolation}
The first benchmark consists of a bias-correction step, followed by bi-linearly interpolating the coarse data to a finer spatial grid.
This two-step procedure is called Bias Correction Spatial Disaggregation (BCSD) \cite{vandal2019intercomparison} and is both simple and prominently used in the downscaling literature. The reanalysis data is first remapped to match the coarse resolution of the climate model grid. The bias correction step then uses quantile mapping applied to the entire spatial domain. To account for seasonal variations, our implementation uses a moving average (with a window size of $25$ days) centered around each calendar day when computing the quantiles. The temporal pooling ensures robust statistics by considering similar days from the seasonal cycle together.

\paragraph{Benchmark 2: Conditioned frame-to-frame diffusion model}
While our approach leverages a score-based approach, i.e., all states/trajectories are generated simultaneously in a non-autoregressive manner, we compare it to a conventional frame-by-frame modeling approach using a denoising diffusion implicit model (DDIM) \cite{ddim2021song}.
We use a time-conditional U-Net backbone, incorporating residual blocks and self-attention layers. The conditioning is then performed by bi-linearly upsampling the bias-corrected climate data to match the high-resolution data and concatenating them along the channel dimension. To predict the middle frame, we condition on a sequence of three frames \citep{schmidt2024wind}.

\paragraph{Metrics}
We use an approximate (sliced) two-dimensional Wasserstein distance \cite{bischoff2024practical} as a metric to compare two high-dimensional probability distributions, defined as
\begin{equation}
    \mathrm{W}(P_{\text{pred}}, P_{\text{ref}}) = \inf_{\gamma \in \Pi(P_{\text{pred}}, P_{\text{ref}})} \mathbb{E}_{(x,y) \sim \gamma} [ \| x - y \|_1 ],
\end{equation}
where $\Pi(P_{\text{pred}}, P_{\text{ref}})$ is the set of couplings, that is, probability distributions whose marginals are $P_{\text{pred}}$ and $P_{\text{ref}}$.
This variant makes the traditional Wasserstein distance computationally feasible for high-dimensional probability distributions \cite{bischoff2024practical}.
In the sliced variant, the high-dimensional data is projected onto a one-dimensional line for which a one-dimensional Wasserstein distance can be efficiently computed. This slicing process is performed repeatedly for multiple different slices, and the result is averaged to obtain a reliable metric.

The Mean Energy Log Ratio (MELR) is used to evaluate the preservation of potentially highly varying physical patterns in the downscaled spatial patches.
The MELR is a metric that is derived from the radially averaged power spectral density (\cref{subsec:supp:rapsd}) as
\begin{equation}
 \mathrm{MELR} = \sum_{k} \left| \log \left( \frac{E_{\text{pred}}(k)}{E_{\text{ref}}(k)} \right) \right|,
\end{equation}
where the energy of the predicted field $E_{\text{pred}}(k)$ is compared to that of the reference field $E_{\text{ref}}(k)$.

We use the structural similarity index measure (SSIM) \citep{wang2003ssim}, which takes into account perceptual properties of local structures when quantifying the similarity between two spatial data points.
The SSIM is defined by sliding a window $W_k(x, y)$ of size $k \times k$ along the spatial data point that computes for two $k\times k$-patches $x$ and $y$
\begin{equation}
    W_k(x, y) = \frac{(2\mu_x\mu_y + c_1)(2\sigma_{x,y} + c_2)}{(\mu_x^2 + \mu_y^2 + c_1)(\sigma_x^2 + \sigma_y^2 + c_2)},
\end{equation}
where $\mu_x, \sigma_x$ and $\mu_y, \sigma_y$ are the respective average and standard-deviations of the values in the respective patches $x$ and $y$. Analogously, $\sigma_{xy}$ is the covariance between the patches.
Finally, $c_1$ and $c_2$ are constants to make the computation more robust.
The SSIM takes on values between $0$ and $1$, whereby higher values indicate more structural similarities between the compared data points. An SSIM of $1$ can only be attained when both data points are identical.
In \cref{tab:supp:metrics} we report the SSIM using a window size of $k = 15$.

\begin{table}[h!]
\centering
\caption{
    \textbf{Quantitative evaluation of downscaling methods.}
    We compare the performance of our model, based on score-based data assimilation (SDA), to a conditional denoising diffusion implicit model (DDIM) and to bias-correction spatial disaggregation (BCSD). We report the sliced Wasserstein-1 distance (Sliced W1) over time, temporally averaged energy log ratio (MELR), and structural similarity index (SSIM) for the four variables: mean sea-level pressure (\texttt{psl}), surface (2m) air temperature (\texttt{tas}), surface (10m) zonal \texttt{uas}, and meridional (\texttt{vas}) wind speeds. The values are reported as mean $\pm$ standard deviation over the generated predictions.
    Note that, since the benchmark methods are not capable of temporal downscaling, the quantitative evaluation is performed on the 6-hourly observation grid.
    Even though the method proposed in this work is under the additional constraint of generating temporally consistent 1-hourly trajectories, it beats both benchmarks in most cases. Between all approaches, the best performance per metric and variable is highlighted in \textbf{bold}.
}
\label{tab:supp:metrics}
\begin{tabular}{ll|l|l|l}
    \toprule
    \diagbox[width=10em]{\textbf{Metric}}{\textbf{Method}} &var&
    \textbf{SDA (ours)} &
    \textbf{BCSD} &
    \textbf{DDIM} \\
    \midrule
    \multirow{4}{*}{Sliced W1 $\downarrow$} & \texttt{psl} & $0.3028 \pm 0.0012$%
                                                           & $\textbf{0.2900} \pm 0$%
                                                           & $0.3526 \pm 0.0536$ \\
                                            & \texttt{tas} & $0.3526 \pm 0.0005$%
                                                           & $0.3999 \pm 0$%
                                                           & $\textbf{0.2894} \pm 0.0999$ \\
                                            & \texttt{uas} & $\textbf{0.1998} \pm 0.0015$%
                                                           & $0.2523 \pm 0$%
                                                           & $0.4055 \pm 0.0586$ \\
                                            & \texttt{vas} & $\textbf{0.2348} \pm 0.0033$%
                                                           & $0.2702 \pm 0$%
                                                           & $0.3553 \pm 0.0371$ \\
    \midrule
    \multirow{4}{*}{MELR $\downarrow$} & \texttt{psl} & $\textbf{1.0256} \pm 0.0031$%
                                                      & $1.1625 \pm 0$%
                                                      & $1.0874 \pm 0.1022$ \\
                                       & \texttt{tas} & $\textbf{0.3470} \pm 0.0018$%
                                                      & $1.1312 \pm 0$%
                                                      & $0.3932 \pm 0.0187$ \\
                                       & \texttt{uas} & $\textbf{1.4622} \pm 0.0025$%
                                                      & $2.4141 \pm 0$%
                                                      & $1.5598 \pm 0.0770$ \\
                                       & \texttt{vas} & $1.4555 \pm 0.0096$%
                                                      & $2.4042 \pm 0$%
                                                      & $\textbf{1.3317} \pm 0.0664$ \\
    \midrule
    \multirow{4}{*}{SSIM $\uparrow$} & \texttt{psl} & $0.8925 \pm 0.0002$%
                                                    & $\textbf{0.9027} \pm 0$%
                                                    & $0.8691 \pm 0.0074$ \\
                                     & \texttt{tas} & $\textbf{0.8524} \pm 0.0002$%
                                                    & $0.7066 \pm 0$%
                                                    & $0.8382 \pm 0.0094$ \\
                                     & \texttt{uas} & $\textbf{0.1351} \pm 0.0008$%
                                                    & $0.0844 \pm 0$%
                                                    & $0.1182 \pm 0.0061$ \\
                                     & \texttt{vas} & $\textbf{0.1322} \pm 0.0009$%
                                                    & $0.0973 \pm 0$%
                                                    & $0.1185 \pm 0.0068$ \\
    \bottomrule
    \end{tabular}
\end{table}

\cleardoublepage

\section{Approximating the gradient of the observation model}\label{subsec:supp:approx-grad}

Using Bayes' rule, we obtain the posterior score function
\begin{align}\label{eq:posterior-score-1}
    \nabla_{X(\difftime)} \left[\log p_\difftime(X(\difftime) \mid y)\right]
    &= \nabla_{X(\difftime)} \left[\log p_\difftime(X(\difftime)) + \log p(Y \mid X(\difftime))\right] \\\label{eq:posterior-score-2}
    &= \underbrace{\nabla_{X(\difftime)} \left[\log p_\difftime(X(\difftime))\right]}_{\approx s_\theta(X(\difftime), \difftime)} + \nabla_{X(\difftime)} \left[\log p(Y \mid X(\difftime))\right].
\end{align}
This illustrates that the conditioning mechanism, which involves only a simple addition of the gradient of the log-observation model, is independent of the trained score model.
However, as detailed by \citet{dps2023chung}, the posterior score requires relating the measurement $Y$ to the diffused state $X(\difftime)$.
\Citet{dps2023chung} propose to approximate
\begin{align}\label{eq:yt-y0-appox-1}
    p(Y \mid X(\difftime)) &= \int p(Y \mid X(0), X(\difftime)) ~ p(X(0) \mid X(\difftime)) ~\dif X(0) \\\label{eq:yt-y0-appox-2}
    &= \int p(Y \mid X(0)) ~ p(X(0) \mid X(\difftime)) ~\dif X(0) \\\label{eq:yt-y0-appox-3}
    &= \mathbb{E}_{X(0) \sim p(X(0) \mid X(\difftime)}\left[p(Y \mid X(0))\right] \\[2mm]\label{eq:yt-y0-appox-4}
    &\approx p\left(Y \mid \hat{X}(0) := \mathbb{E}_{X(0) \sim p(X(0) \mid X(\difftime)}\left[X(0)\right]\right).
\end{align}
Intuitively, the last step pulls the expectation into the conditioning, which is not equivalent in general, and thereby approximates $p(Y \mid X(\difftime)) \approx p(Y \mid \hat{X}(0))$. The quantity $\hat{X}(0)$ is a posterior-mean estimate for the noise-free data point underlying the diffused state $X(\difftime)$ at the current diffusion-time $\difftime$ in the generative process.
In the following, we detail how $\hat{X}(0)$ is computed.

Linear stochastic differential equations, like the diffusion process defined in \cref{eq:forward-diffusion-process}, can be equivalently formulated in terms of a discrete, linear Gaussian transition model
\begin{equation}\label{eq:discrete-forward-process}
    p(X(\difftime + \Delta \difftime) \mid X(\difftime)) = \mathcal{N}\left(X(\difftime + \Delta\difftime); \mat{A}(\Delta \difftime)X(\difftime), \mat{\Sigma}(\Delta \difftime)\right),
\end{equation}
for some time increment $\Delta \difftime$.
We refer to, e.g., \citet[Section 6.1]{sdebook2019sarkka} for more details on how to derive the \emph{transition} and \emph{process-noise covariance} functions $\mat{A}$ and $\mat{\Sigma}$ from the drift $\mat{F}$ and dispersion $\mat{L}$ of the diffusion process (\cref{eq:forward-diffusion-process}).
The forward diffusion process can be simulated over extended time ranges in a single step via sampling from the Gaussian transition density
\begin{equation}\label{eq:discrete-forward-process-add}
    X(\Delta \difftime) \mid X(0) = \mat{A}(\Delta \difftime)X(0) + \mat{\Sigma}^{\frac{1}{2}}(\Delta \difftime)\epsilon, \quad \epsilon \sim \mathcal{N}(0, \idmat),
\end{equation}
which is an equivalent formulation of \cref{eq:discrete-forward-process} for the case $\difftime = 0$, i.e., starting from the noise-free data point \citep{ddpm2020ho,sdedm2021song,ddim2021song}.
The same is not possible in the reverse direction, which would make the generation process trivial. Unfortunately, the backwards transition model associated with \cref{eq:backward-diffusion-process} is Gaussian only for infinitesimally small time decrements $\Delta \difftime \rightarrow 0$ \citep{feller1949theory,dm2015sohldickstein}, which leads to sampling quality increasing with the number of steps used for simulating the generative process.
However, it is possible to estimate the posterior mean $\hat{X}(0)$ based on the current score estimate $s_\theta(X(\difftime), \difftime)$. By re-arranging \cref{eq:discrete-forward-process-add} we note that
\begin{align}\label{eq:xhat-1}
    X(0)
    &= \mat{A}^{-1}(\difftime)\left(X(\difftime) - \mat{\Sigma}^{\frac{1}{2}}(\difftime)\epsilon\right) \\\label{eq:xhat-2}
    &= \mat{A}^{-1}(\difftime)\left(X(\difftime) + \nabla_{X(\difftime)}\left[p_\difftime(X(\difftime))\right]\mat{\Sigma}(\difftime)\right) \\\label{eq:xhat-3}
    &\approx \mat{A}^{-1}(\difftime)\left(X(\difftime) + s_\theta(X(\difftime), \difftime)\mat{\Sigma}(\difftime)\right) \\\label{eq:xhat-4}
    &= \hat{X}(0).
\end{align}
The first equality uses the direct correspondence between $\epsilon$ and the de-noising score function of \cref{eq:discrete-forward-process} \citep{ddpm2020ho,variationaldms2021kingma}.
Since the true score, like $\epsilon$, is not known while generating a new data point, it is in the next step approximated with our parametric score model $s_\theta$.
A more detailed derivation can be found for the scalar case ($A, \Sigma \in \mathbb{R}$) in \citet{dps2023chung}.

With that, we established the necessary background regarding the conditional model, in order to describe the approximation to the conditioning mechanism that is used in this work.
Plugging into \cref{eq:posterior-score-1,eq:posterior-score-2} the approximation from \crefrange{eq:yt-y0-appox-1}{eq:yt-y0-appox-4} reveals that we have to compute the gradient
\begin{equation}
    \nabla_{X(\difftime)}\left[\log p(Y \mid \hat{X}(0))\right].
\end{equation}
For our purposes, we assume a Gaussian observation model (\cref{eq:observation-model}), which simplifies this gradient to
\begin{align}
    \nabla_{X(\difftime)}\left[\log p(Y \mid \hat{X}(0))\right]
    &= \nabla_{X(\difftime)}\left[\log \mathcal{N}\left(Y; h(\hat{X}(0)), \mat{R}\right)\right]\\
    &= \nabla_{X(\difftime)}\left[\lVert Y - h(\hat{X}(0))\rVert_\mat{R}\right],
\end{align}
where $\lVert\cdot\rVert_\mat{R}$ is a Mahalanobis norm in the multivariate case and a Euclidean norm in the scalar case.
Let us denote the residual between conditioning information and the observed predicted signal as $r(Y, \hat{X}(0)) := Y - h(\hat{X}(0))$.
Then, using the chain rule of differentiation, we obtain
\begin{equation}
    \nabla_{X(\difftime)}\left[\lVert r(Y, \hat{X}(0))\rVert_\mat{R}\right]
    = \nabla_{r(Y, \hat{X}(0))}\left[\lVert r(Y, \hat{X}(0))\rVert_\mat{R}\right] \cdot \nabla_{\hat{X}(0)}\left[h\left(\hat{X}(0)\right)\right] \cdot \nabla_{X(\difftime)}\left[\hat{X}(0)\right].
\end{equation}
We plug the equality from \cref{eq:xhat-3,eq:xhat-4} into the third component of the gradient to obtain
\begin{align}
    \nabla_{X(\difftime)}\left[\hat{X}(0)\right]
    &= \nabla_{X(\difftime)}\left[\mat{A}^{-1}(\difftime)\left(X(\difftime) + s_\theta(X(\difftime), \difftime)\mat{\Sigma}(\difftime)\right)\right] \\
    &= \mat{A}^{-1} (1 + \nabla_{X(\difftime)}\left[s_\theta(X(\difftime), \difftime)\mat{\Sigma}(\difftime)\right] \\
    &\approx \mat{A}^{-1}.
\end{align}
The final step approximation avoids the computational demanding differentiation of the score function with respect to the perturbed state, which cuts away the majority of the computational cost in the conditioning.
Altogether, the approximate conditioning term (c.f.~\cref{eq:posterior-score-2}), used throughout this work, is
\begin{equation}
    \begin{aligned}
    \nabla_{X(\difftime)}\left[\log p(Y \mid X(\difftime)\right]
    &\approx \nabla_{X(\difftime)}\left[\log p(Y \mid \hat{X}(0))\right]\\
    &\approx \nabla_{r(Y, \hat{X}(0))}\left[\lVert r(Y, \hat{X}(0))\rVert_\mat{R}\right] \cdot \nabla_{\hat{X}(0)}\left[h\left(\hat{X}(0)\right)\right] \cdot \mat{A}^{-1},
    \end{aligned}
\end{equation}
which is very cheap to compute in each de-noising step.

\section{Anomalies}

\begin{figure}[h!]
    \centering
    \includegraphics[width=\linewidth]{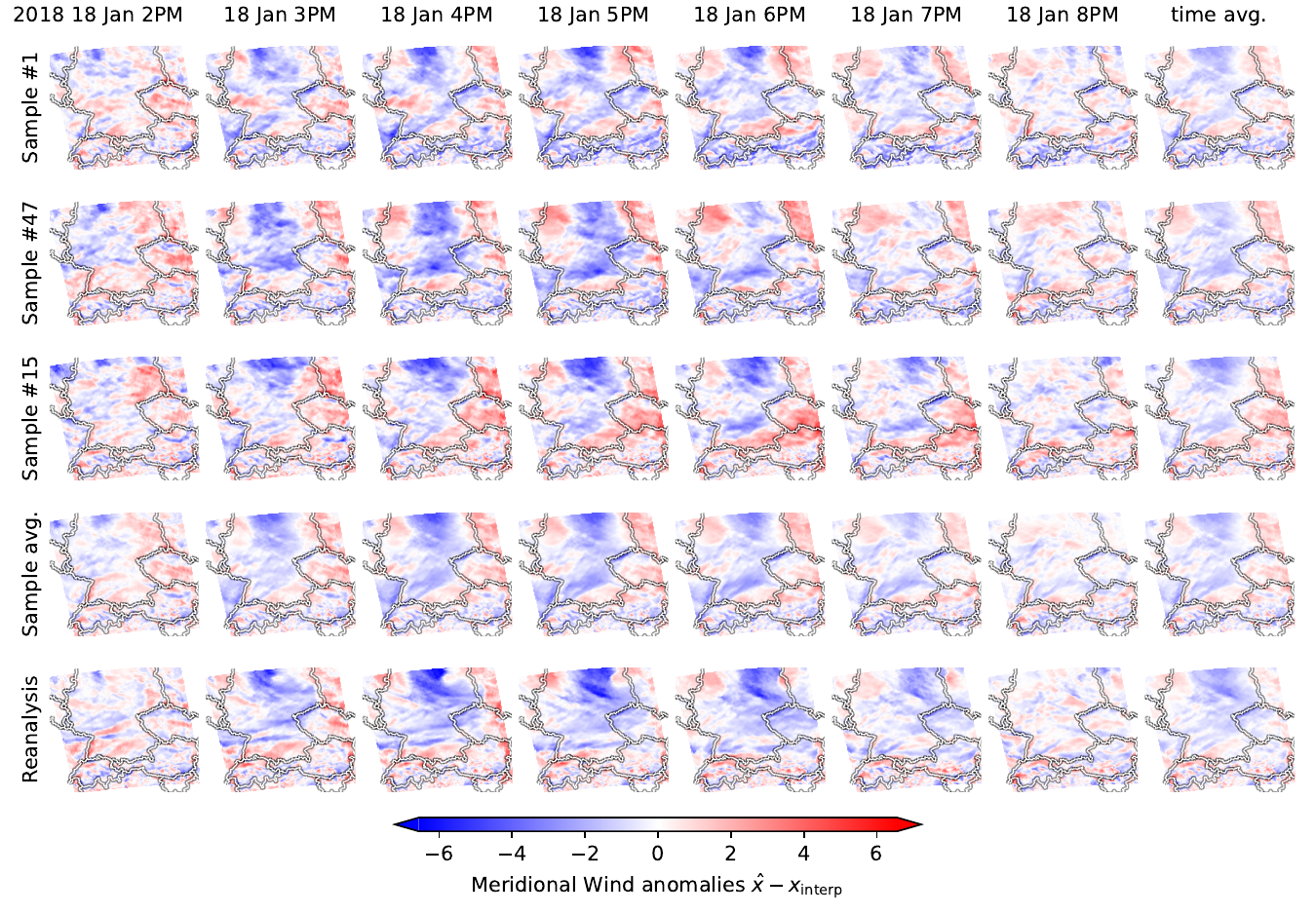}
    \caption{\textbf{Anomalies between downscaled and interpolated versions of coarse input.}
    This plot supplements Figure 4. It visualizes the differences between a spatiotemporal interpolation of the coarse input $x_\mathrm{interp}$ (Figure 4, fourth row) and different fine-scale time series $\hat{x}$: 1) three samples from the proposed downscaling model (\emph{top three rows}), 2) the average of the three samples (\emph{fourth row}), and 3) the ground-truth reanalysis data (\emph{bottom row}).
    This visualization exposes the local spatial and temporal patterns on the fine grid that are not contained in the coarse data.
    The rightmost column plots the corresponding temporal averages of the spatial anomalies.
    As in Figure 4, the downscaling model is only conditioned at 2PM (first column) and 8PM (penultimate column).
    Especially at those conditioning points, the local spatial patterns predicted in each sample (rows 1 through 3) are structurally similar to those in the reanalysis data (bottom row).
    Between the conditioning points (3PM through 7PM), the ground-truth anomalies (bottom row) expose that the temporal evolution of the cyclone is not predicted by the smooth temporal interpolation.
    Our model predicts spatiotemporal structure on the fine grid, adding information that cannot be trivially inferred from the coarse input.
    }
    \label{fig:supp:storm-anomalies}
\end{figure}

\clearpage

\section{Generative de-noising process}

\begin{figure}[h!]
    \centering
    \includegraphics[width=\linewidth]{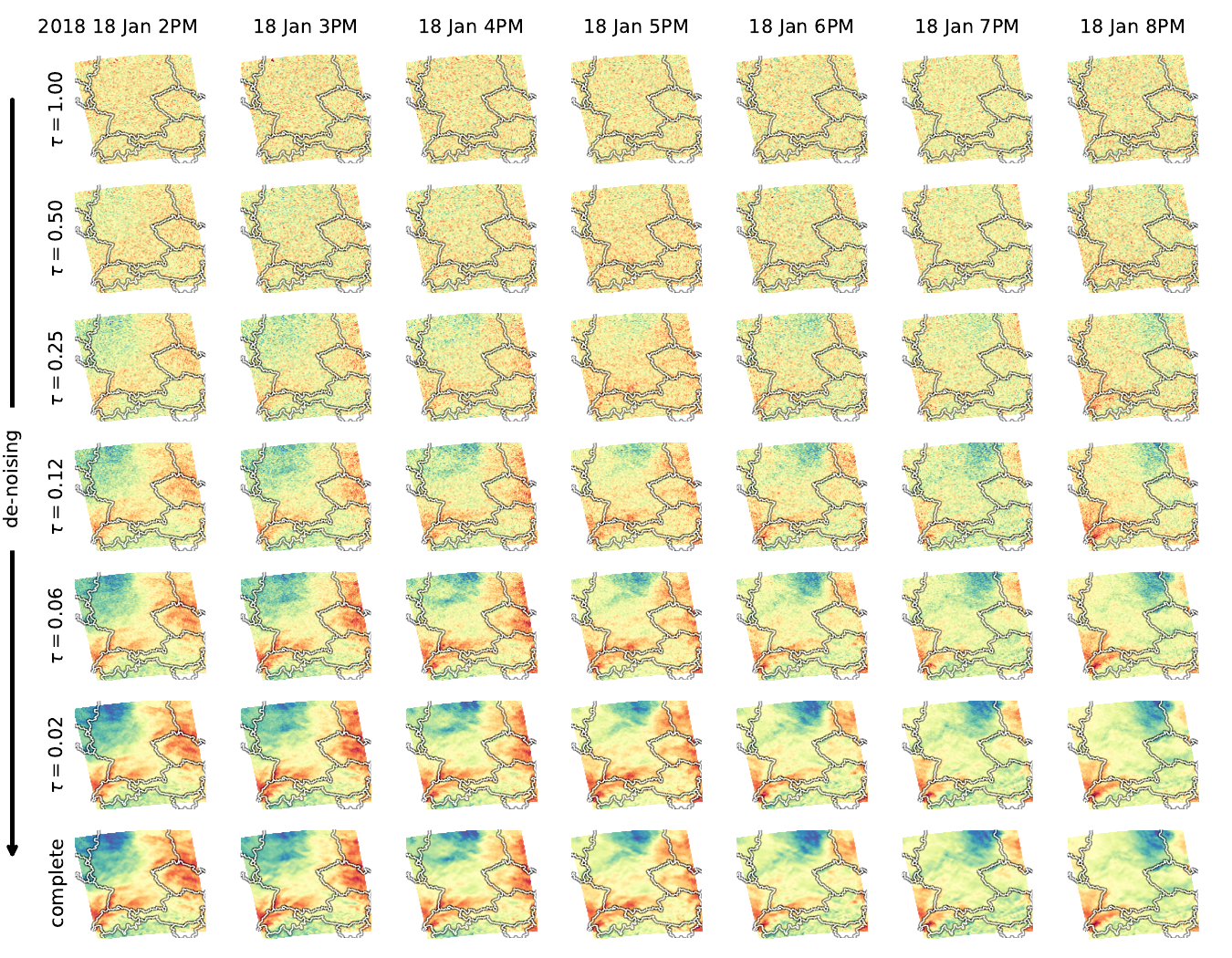}
    \caption{\textbf{The generative process of the diffusion model.} Diffusion models learn a mapping from a tractable noise distribution ($\difftime = 1$; often Gaussian noise) to the training-data distribution.
    Using a statistical model for the score function, which separates noise from signal, an initial random-noise sample is iteratively de-noised into a data point that lies in a region of high data density.
    This figure shows seven (of 256) steps of this generative process, beginning at $\difftime = 1$ (\emph{top row}; Gaussian noise) and ending at $\difftime=0$ (\emph{bottom row}).
    Fine-scale features are generated by the model towards the end of the generative process.
    Notably, the spatial and temporal structure emerge jointly, since the score function is estimated for the entire time series.
    }
    \label{fig:supp:denoising}
\end{figure}

\clearpage

\section{Relationship between downscaled variables}

This experiment demonstrates that the prediction for one variable is affected by conditioning information about the other remaining variables through inter-variable relationships that the generative model learned through training.
To show this, we isolate one variable "v" of interest (here: meridional wind speeds) and denote the remaining variables (here: mean sea-level pressure, surface temperature, zonal wind speeds) as "$\neg$v".
We predict four different downscaled sequences for v:
\begin{enumerate}
    \item First, we draw from the prior. Sampling from the unconditioned generative downscaling model, yields an uninformed sequence of weather patterns.
    \item Second, we draw from the generative model that is conditioned only on $\neg$v.
    \item Third, we draw from the generative model that is conditioned only on v.
    \item Finally, we draw from the fully conditioned generative downscaling model, providing the model with the information about all considered variables, v and $\neg$v, as is the default case in the other experiments (e.g., Figure 4).
\end{enumerate}

\begin{figure}
    \centering
    \includegraphics[width=\linewidth]{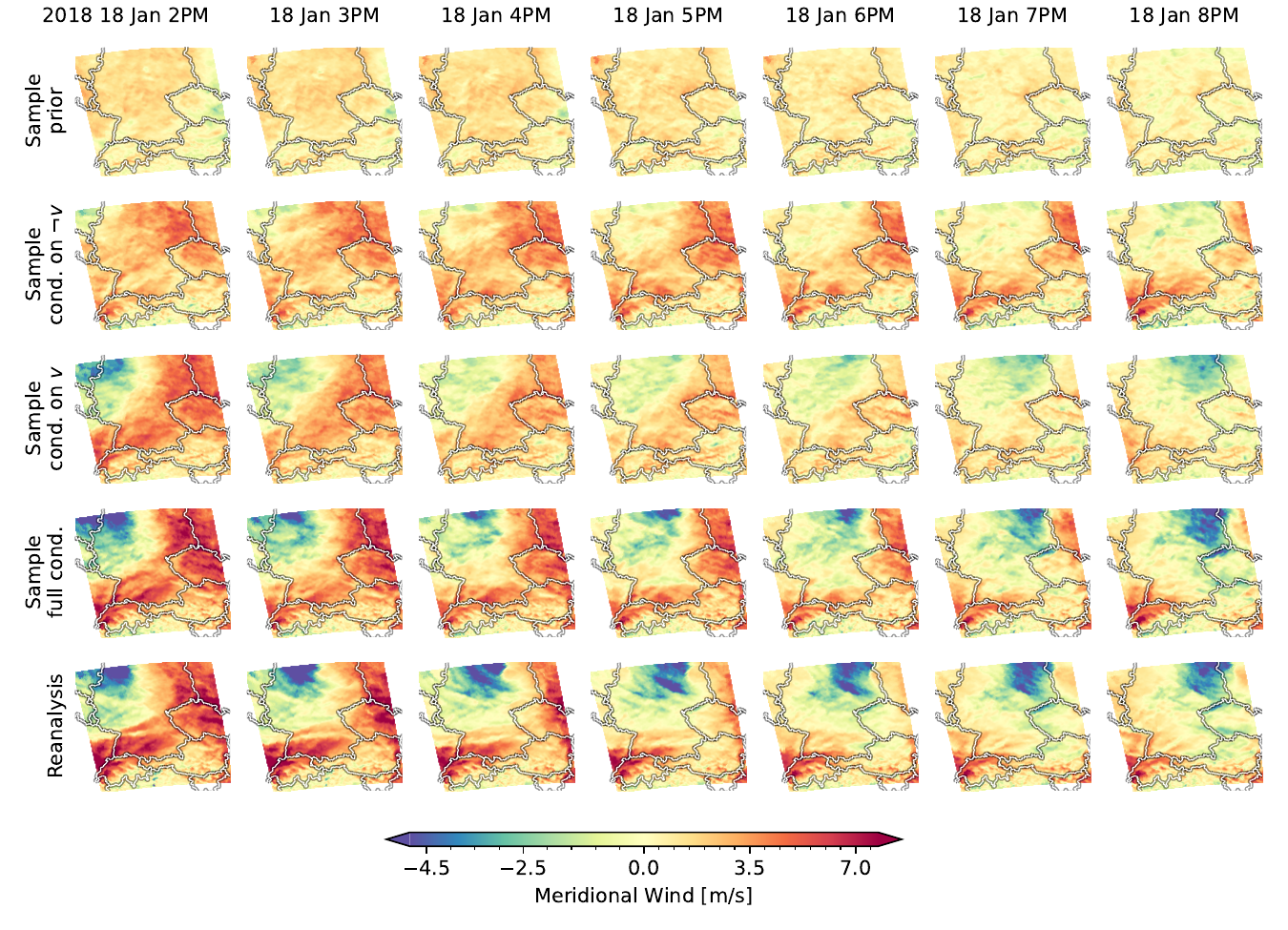}
    \caption{\textbf{The model learns and uses relationships between variables.}
    The first four rows of this plot visualize four different predictions for downscaled meridional wind during a cyclone ("Friederike, January 2018).
    The bottom row shows the reanalysis data for comparison.
    For the conditioned predictions (rows 2 through 4), no information is provided to the model between the first (2:00 PM) and the last (8:00 PM) visualized time point, exactly as in Figure 4.
    We denote the visualized variable of interest (here: meridional wind speeds) as "v", and the other variables (here: mean sea-level pressure, surface temperature, and zonal wind speeds) as "$\neg$v".
    As in Figure 4, the sign of the wind speed value defines its direction and time progresses from left to right hourly, starting 2018 January 18 at 02:00 PM and ending the same day at 08:00 PM.
    The \emph{first row} shows a sample from the unconditioned generative model, which is entirely uninformed by any coarse input.
    The \emph{second row} shows the downscaled v, predicted by the generative model that is only conditioned on $\neg$v.
    The \emph{third row} shows the downscaled v, predicted by the generative model that is only conditioned on v.
    The \emph{fourth row} shows the downscaled v, predicted by the generative model conditioned on all variables, v and $\neg$v.
    The plot serves to demonstrate the effect of the multivariate nature of the downscaling model.
    Comparing the model outputs when conditioning on different sets of variables demonstrates that the prediction of a variable v is affected by incorporating information about the other variables $\neg$v.
    This allows the conclusion that the downscaling model has learned inter-variable relationships, which it uses for prediction.
    }
    \label{fig:supp-othervar}
\end{figure}

\cref{fig:supp-othervar} visualizes these differently-informed samples of v as spatiotemporal sequences (cf.~Figure 4).
A comparison between the unconditioned sample (1.; first row in \cref{fig:supp-othervar}), the different partly-conditioned predictions (2. and 3.; second and third rows in \cref{fig:supp-othervar}), and the fully-conditioned predictions (4.; fourth row in \cref{fig:supp-othervar}), demonstrates that the multivariate downscaling model has learned relationships between the variables, which it uses for generating downscaled predictions.
Note that this experiment is purely diagnostic and serves to validate that the model learns and does not neglect relationships between the considered variables.

\end{document}